\documentclass{article}
\usepackage{enumitem}

\usepackage[preprint]{neurips_2022}

\newcommand{\resbn}{\textit{resnet18\_bn} }
\newcommand{\resgn}{\textit{resnet18\_gn} }
\usepackage[utf8]{inputenc} %
\usepackage[T1]{fontenc}    %
\usepackage{xcolor}         %
\usepackage[font=small]{caption}
\usepackage{microtype,booktabs,graphicx,subcaption}
\usepackage{amsmath,amssymb,amsthm,amsfonts,mathtools}
\usepackage{multirow}
\usepackage{rotating}

\usepackage{physics}

\usepackage{array}

\usepackage{hyperref}
\usepackage{url}            %

\usepackage{thmtools,thm-restate,amsthm}
\usepackage{verbatim}

\usepackage{ragged2e}%
\usepackage{natbib}

\newcommand{\remove}[1]{{}}

\newcommand{\ie}{\textit{i.e., }}
\newcommand{\eg}{\textit{e.g., }}

\title{Generalization to translation shifts: a study in architectures and augmentations}

\author{Suriya Gunasekar \\ Microsoft Research \\ \texttt{suriya@ttic.edu}}

\begin{document}

\maketitle

\begin{abstract}

We study how effective data augmentation is at capturing the inductive bias of carefully designed network architectures  for spatial translation invariance. We evaluate various image classification architectures (antialiased, convolutional, vision transformer, and fully connected MLP networks) and data augmentation techniques towards generalization to large translation shifts. We observe that: (a) without data augmentation, all architectures, including convolutional networks with antialiased modification suffer some degradation in performance when evaluated on translated test distributions. Understandably, both the in-distribution accuracy and degradation to shifts is significantly worse for non-convolutional models.  (b) The robustness of performance is improved by even a minimal augmentation of $4$ pixel random crop across all architectures. In some instances, even $1-2$ pixel random crop is sufficient. This suggests that there is a form of meta generalization from augmentation. For non-convolutional architectures, while the absolute accuracy is still low with this basic augmentation, we see substantial improvements in robustness to translation shifts. (c) With a sufficiently advanced augmentation pipeline ($4$ pixel crop+RandAugmentation+Erasing+MixUp), all architectures can be trained to have competitive performance in terms of in-distribution accuracy as well as generalization to large translation shifts.
\end{abstract}

\section{Introduction}\label{sec:intro}

Convolutional neural networks (ConvNets) are a natural architectural choice for a variety of computer vision tasks. The built-in structure from localization and translation equivariance of the convolutional layers is intrinsically useful in many image processing scenarios \citep{krizhevsky2012imagenet,lecun1989backpropagation,fukushima1982neocognitron}. For over a decade ConvNets were the backbone of computer vision and continue to be one of the most important class of models. At the same time, advances in large datasets and data augmentation techniques have made it possible to train  general purpose  architectures to be competitive with ConvNets despite lacking any image specific priors. Most popular among these are the Vision Transformers (ViTs) and their variants  \citep{vaswani2017attention,dosovitskiy2020image,touvron2021going,touvron2021training}. When pretrained on ultra-large datasets like ImageNet-21k (14 million images) or JFT-300/3B (300 million/3 billion weakly labeled images, respectively), ViTs can outperform similarly pretrained ConvNets on diverse vision tasks. While the scale of the data was original thought to be crucial,  follow up work show that competitive accuracies can  also be achieved in small-to-medium data regimes using advanced data augmentation \citep{touvron2021training} or optimization techniques \citep{chen2021vision}. 
Detailed experimentation on competitive benchmarks by \citet{steiner2021train} showed that ViTs trained with extensive data augmentation can recover the performance gains from $\sim10$x larger independently annotated dataset.   
In such large data or extensive augmentation regimes, even simpler fully connected multi-layer perceptron (MLP) architectures \citep{tolstikhin2021mlp,touvron2021resmlp} can achieve competitive performance. In this work we focus on the role of data augmentation in learning from general purpose architectures. %

\textit{Can data augmentation capture the inductive biases of carefully designed architectures?} %
Beyond accuracy, the design of architectures is often motivated by domain knowledge of desired invariances. One of the  fundamental  image priors is the invariance of object labels to spatial shift or translation of its position. Indeed, the success and motivation of ConvNets is often attributed to their component convolutional operators being definitionally translation equivariant. In this work, we study to what extent the  shift-equivariant convolutional layers make ConvNets robust to translation shifts compared to ViTs and MLPs? How effective are  data augmentation techniques in encouraging similar behaviors? 

There is a rich literature on understanding the translation invariance properties of ConvNets (see Section~\ref{sec:related}). It has been shown that despite the equivariance of the convolutional operator, other architectural components like non-linearities and strides can cause the networks lose their invariance to translations \citep{zhang2019making,azulay2019deep,chaman2021truly,engstrom2019exploring,xiao2018spatially,alsallakh2020mind}. Despite these findings, it is reasonable to believe that ConvNets will still have substantial edge in robustness to translation when compared to ViT or MLP models. At the same time, the competitive performance of latter architectures suggests that good image priors can also be learned from rich training data. The goal of this work is to quantify the relative effectiveness  of architectures design and data augmentation towards robustness to translation shifts. Algorithmically, these are complementary tools for incorporating domain knowledge. On one hand, it is conceptually simpler to generate data with desired invariant transformation rather than hard code it in the architecture. On the other hand, data augmentation only provides a weak supervision about the invariant properties and is limited by the biases  in the training samples.

\paragraph{Generalization vs invariance.} In our experiments, rather than chase the gold standard of \textit{strict} invariance on all inputs, we work with a data-centric measure of \textit{generalization to translation shifts} in the test distribution. We evaluate our models for accuracy on specific out-of-distribution test datasets  where the object locations are systematically shifted without creating additional distortions or domain gaps from training distribution.  See Section~\ref{sec:expsetup} for full  setup.

In the strict sense, a classifier $f:X\to Y$ on input space $X$  is invariant to set of spatial translations $T$ if for all $t\in T$ and all inputs $x\in X$, $f(t(x))\approx f(x)$. %
Often we are not concerned with performance on all inputs but rather on typical samples from a task distribution, say $(x,y)\sim D$. We thus evaluate our models a structured out-of-distribution accuracy, wherein we train on samples from $D$, but test on translation shifted inputs $t(x)$, \ie generalized accuracy to  shift $t$ is %
    $\mathbb{E}_{(x,y)\sim D} \mathbf{1}[f(t(x))= y]$.
This evaluation also differs from adversarial robustness to translation shifts as studied in \citet{engstrom2019exploring,xiao2018spatially}. In our notation, the adversarial accuracy metric would be $\mathbb{E}_{(x,y)\sim D} \mathbf{1}[f(t_x(x))\neq y]$, where $t_x\in T$ is adversarially chosen per-sample $(x,y)$. 
Finally, we clarify that data augmentation does change the training distribution $D$. If inputs $x$ were to be augmented with their respective transformations $\{t(x):t\in T\}$ (and no other augmentation), then there is no distribution shift. However, our experiments are never in the no-distribution shift regime. Importantly, all our augmentation pipelines uses random crop of at most $4$ pixels (some are further restricted to $1$ or $2$ pixels), but we evaluate our models on much larger translation shifts of up to $8$ pixels on CIFAR and $16$ pixels on TinyImageNet. %

\subsection{Related work}\label{sec:related}
\paragraph{Translation invariance in ConvNets. } ConvNets have been extensively studied on various approximate measures of invariance to translation shifts  including, but not limited to, \cite{goodfellow2009measuring, zeiler2014visualizing, fawzi2015manitest, kanbak2018geometric, azulay2019deep, zhang2019making, engstrom2019exploring, kayhan2020translation, chaman2021truly}.   For example, \citet{zeiler2014visualizing} provide visualizations of the hidden layer filters in early ConvNets that show their sensitivity to small changes in translation, scale, and rotation. Similar visual evaluation was also more recently used in \citet{alsallakh2020mind}. Other work \cite{fawzi2015manitest, kanbak2018geometric, azulay2019deep, zhang2019making, engstrom2019exploring} focus on more quantitative measures of invariance such as mean change in top-$1$ prediction or class probabilities between images within a distortion range.  These works collectively establish that despite the built-in inductive bias, ConvNets are not ``truly" invariant even to small translation.  Some recent work, importantly \citep{zhang2019making,chaman2021truly}, address this shortcoming of ConvNets by designing new architectural modifications to ConvNets. %
These prior works show that strict spatial invariance is a strong measure that even ConvNets with their built-in priors do not satisfy. We choose a data-centric measure in our evaluation that is more practical for general  architectures. 
Our experimental and evaluation protocol has some key differences from these prior works studying invariance: (a) Our evaluation is designed to introduce  translations shifts in the test data without with any additional confounding domain gaps between train and test distributions. For example, in the experiments of \cite{azulay2019deep,zhang2019making} the test samples are padded and resized during evaluation, while the models were trained on regular dataset without such padding or resizing. This leads the test dataset to differ from the training data in ways that are not just translation of the objects. (b) Our evaluation is non-adversarial in that it measures degradation in average test loss and not in the worst-case drop in performance on any single image. Further, by comparing the drop in performance from  translation shift to the unperturbed in-distribution accuracy, we inherently down-weight the non-robustness on hard-to-learn inputs on which the classifier had inaccurate prediction to begin with. %
(c) Finally, our evaluation does not penalize models from learning position dependent features as observed by \cite{kayhan2020translation}.  For example, a network has the flexibility to  use its large representation power to create a separate model for an object (say a cat) at each pixel location --- while such a model will be inefficient, it would still do a good job at detecting cats in translated test distribution.  %
These differences are nuanced but significant, which makes our results complementary to prior work in this space. %

\paragraph{Robustness under adversarial perturbations and other OOD benchmarks.} In a work closest to ours, \cite{engstrom2019exploring} study test accuracy degradation from adversarially chosen translations and rotations on test images.  While much of the work in adversarial robustness focus on $\ell_\infty$ or $\ell_2$ norm bounded perturbations to the inputs, \cite{engstrom2019exploring} show that ConvNets can be effectively ``attacked" even when the perturbed inputs are merely small rotations and/or translations of the input. Similar adversarial attack on ConvNets based on  more complex spatial transformations was previously studied in \citep{xiao2018spatially}. Our evaluation is closer to ``random perturbation" evaluation, which is only briefly explored in \cite{engstrom2019exploring}. In comparison to these studies, our distribution shift is picked non-adversarially and independent of input samples. 

Complementing the work on adversarial robustness, there has also been lot of interest in evaluating models on other out-of-distribution robustness.  Generalization to out-of-distribution test datasets is a broad umbrella topic, and aside from adversarial robustness, many work also focus on robustness performance on benchmarks for benign ``natural'' perturbations \cite{recht2019imagenet,hendrycks2018benchmarking,koh2021wilds,djolonga2021robustness}. For ConvNets, inspired by the neural networks \textit{scaling laws} line of work, \cite{djolonga2021robustness} probe for relation between OOD robustness and learning choices like data size, model size, optimization algorithm, and model choices like model sizes, and normalization (they do not consider data augmentation in detail though). They also propose a synthetic benchmark SI-SCORE for controlled image invariance evaluation. Expanding on this line of work, \cite{yung2021si,Bhojanapalli_2021_ICCV, bai2021transformers, mahmood2021robustness,shao2021adversarial,paul2021vision,pinto2022impartial}, compare ConvNets and ViTs on adversarial robustness and/or out-of-distribution robustness benchmarks. %

Aside from the methodological differences, much of the work in this space has focused on the relative merits, demerits, and robustness of different model choices and the role of training data size. The effects of data augmentation is only minimally considered, if at all. In contrast, our goal in this work is to specifically quantify how much different data augmentation pipelines can capture the inductive biases in a carefully designed architecture.

\section{Experimental setup}\label{sec:expsetup}

All our experiments are conducted on the CIFAR-10, CIFAR-100, and TinyImageNet datasets.  
Since our study involves training from scratch and testing on large models in numerous configurations, it is beyond the scope of the paper to extend such a detailed study to larger benchmarks like full ImageNet. Moreover, we emphasize that our goal here is not to get the state-of-the-art accuracy/robustness on benchmarks, but rather to understand how much data augmentation captures the benefits of the convolutional architecture. Arguably, it is also the small data regime where the inductive biases from architecture and/or augmentations play more important roles. In ultra-large scale datasets,  accuracy/robustness might naturally come from dataset size itself rather than model priors. %

\begin{figure}[thb]
\centering
    \begin{minipage}{0.85\linewidth}
    \centering
    \begin{subfigure}[b]{0.16\textwidth}
         \centering
         \includegraphics[width=\textwidth]{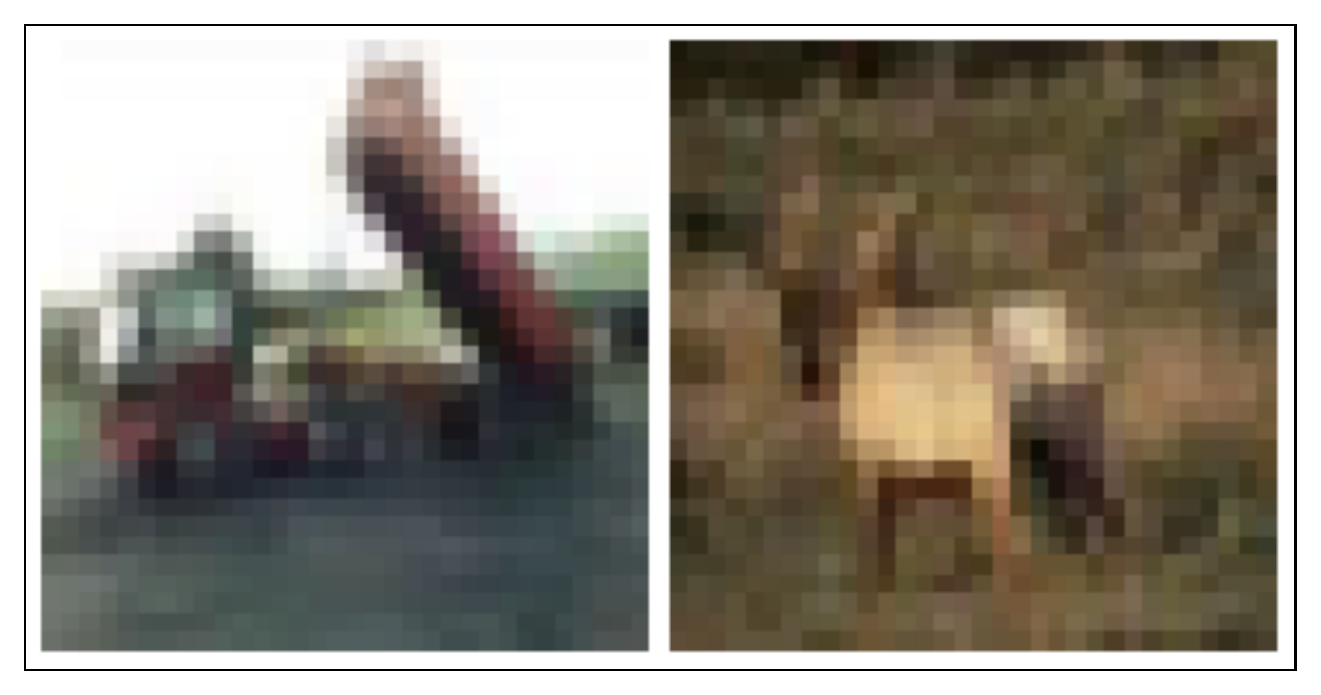}
         \caption[\footnotesize]{Original}
         \label{fig:orig}
     \end{subfigure}~
     \begin{subfigure}[b]{0.24\textwidth}
         \centering
         \includegraphics[width=\textwidth]{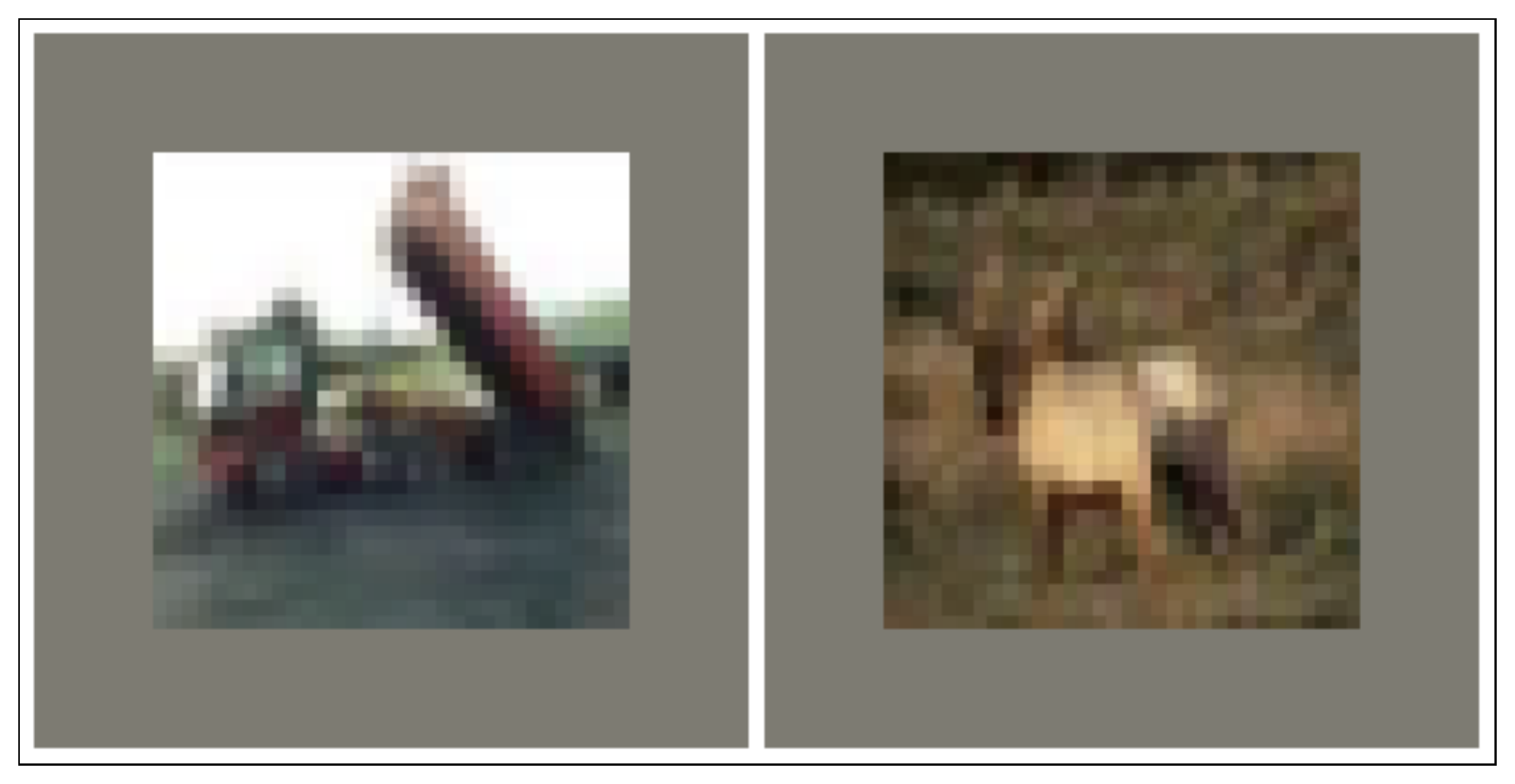}
         \caption[\footnotesize]{Mean padding}
         \label{fig:pad}
     \end{subfigure}~
     \begin{subfigure}[b]{0.24\textwidth}
         \centering
         \includegraphics[width=\textwidth]{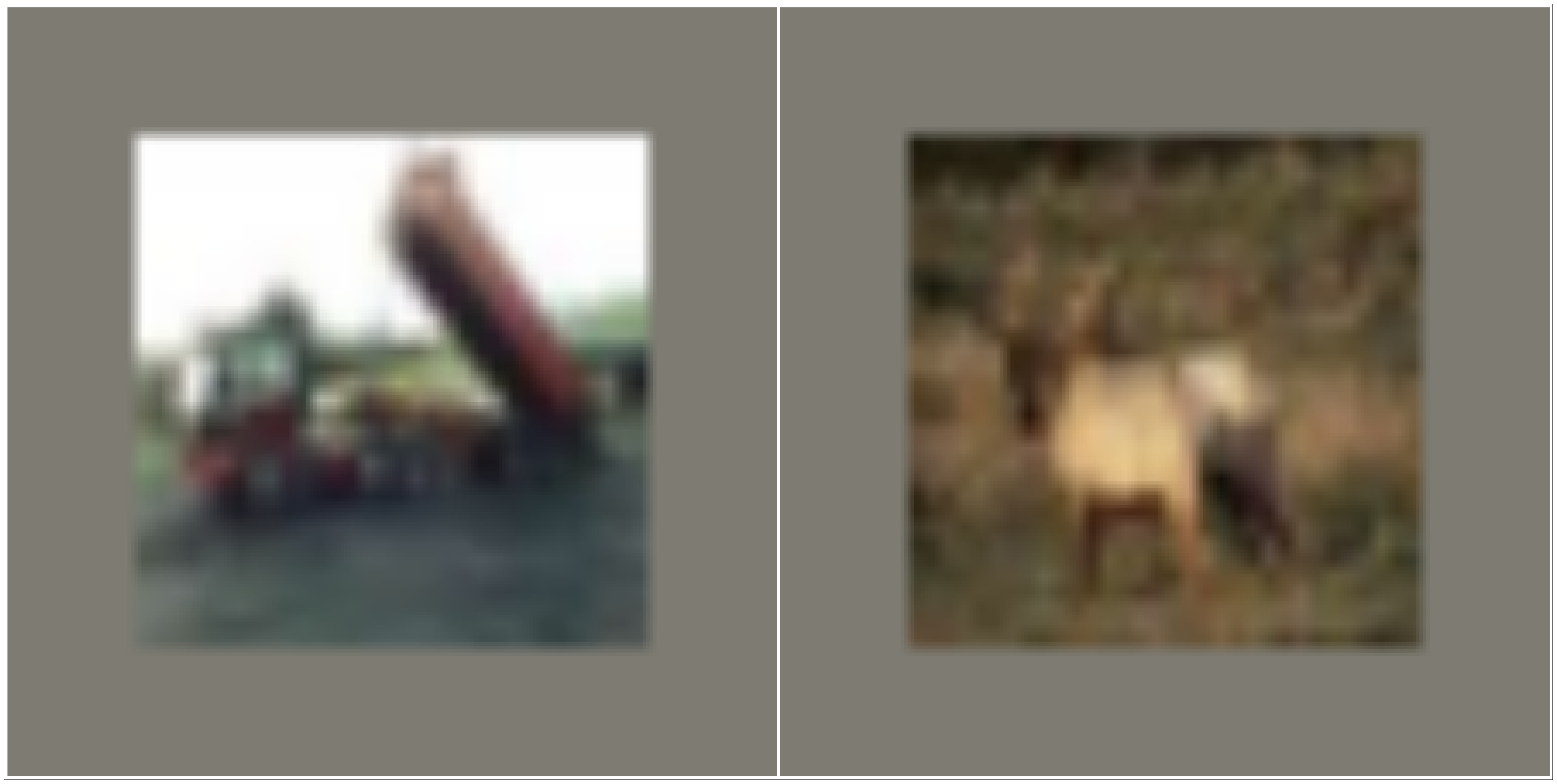}
         \caption[\footnotesize]{Resize to $224\times 224$}
         \label{fig:pad-resize}
     \end{subfigure}~
  \begin{subfigure}[b]{0.24\textwidth}
         \centering
         \includegraphics[width=\textwidth]{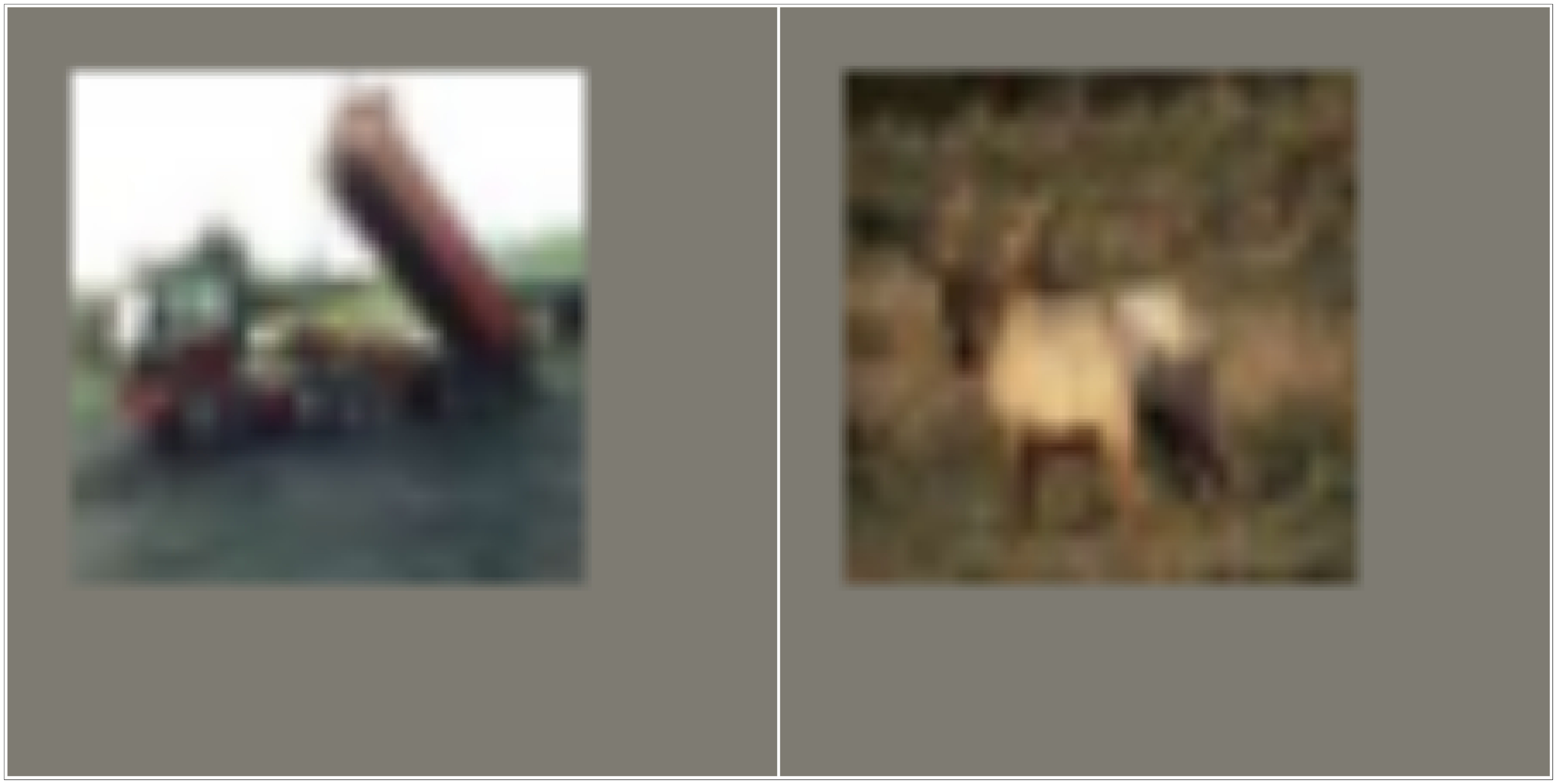}
         \caption[\footnotesize]{Shifted test sample}
         \label{fig:pad-resize-translate}
 \end{subfigure}
 \end{minipage}
\caption{Sample images from the prepossessing steps}
\label{fig:samples}
\end{figure}

 In the main paper we focus on results from CIFAR-10 and CIFAR-100 datasets which consists of $32\times 32$ pixel RGB images balanced across $10$ and $100$ classes, respectively. We defer the discussion on TinyImageNet, which a subset of the more diverse ImageNet benchmark, to Appendix~\ref{sec:tinyimagenet}. 
To study large translation shifts without introducing domain gaps, we modify the  dataset by symmetrically padding all the CIFAR images with $8$ pixels ($1/4$ of image size) on each side. The padded pixels contain the mean channel values of the entire  training dataset, which ensures that the channel-wise means and standard deviations across training dataset remains the same as the original un-padded dataset (see illustration in Figure~\ref{fig:samples}).  This padded dataset allows us to evaluate  large translation shifts of up to $16$ pixels (Hamming distance) in the test dataset without creating additional confounding factors. \textit{Importantly, in all the shifted test sets, there is no cropping or loss of the image content and the entire image is available to the network at the same scale as seen during training.} After padding with a mean-valued canvas, we resize the resulting $48\times 48\times 3$ images to  $224\times224\times 3$ (the standard input size for ImageNet) using bilinear interpolation. This up-sampling step helps avoid extensive hyperparameter tuning of the models, especially, the ViT and MLP models. %

We briefly discuss the alternative evaluation methodologies. The more natural random cropping of images to evaluate robustness to translation is inherently limited by the number of pixels we can faithfully forgo without losing information and hence cannot capture large translation shifts. %
In prior work, \citet{azulay2019deep} also used similar padded images to investigate translation shifts. A key difference in our methodology is that we have our entire training and testing pipeline on the preprocessed images (with padding), while the latter paper evaluated models pretrained on standard ImageNet without any padding -- this creates an uncontrolled distribution shifts.  Another technical difference is that \citet{azulay2019deep} downsampled the images, which leads to loss in resolution, while our preprocessing is non-lossy. Finally, synthetic benchmarks such as SI-Score proposed in \cite{djolonga2021robustness} are a good alternative to our setup. However, for our simple controlled setting, the conceptual advantage of padding is that it does not change the natural distribution of foreground and background which would be lost in the cut-and-paste protocol of \cite{djolonga2021robustness}. Furthermore, the segmentation process  in \cite{djolonga2021robustness} appears to be not perfect which might create additional confounders. To further ensure that our ``synthetic" padding is benign,  we verified that the test accuracy of our models trained on the padded dataset is comparable to the standard  train-test pipeline on $32\times32\times 3$ CIFAR inputs without any padding (see Table~\ref{tab:in-distribution}). 

\paragraph{Training} All the models are trained on the mean-padded  datasets for $1600$ epochs on $8\times V100$ GPUs. We use implementations (with suitable modifications) of the models from various open source repositories, most notably \citet{rw2019timm} and \citet{liu-github}. We performed basic hyperparameter tuning in a small grid around the parameters reported in the respective papers. The exact value of hyperparameters used in experiments along with code are provided in the supplementary material. All the evaluation metrics reported in this paper are median performance over $3$ runs. 

\subsection{Architectures}
Our goal is to compare fundamentally different architectures on  generalization to translation shifts. 
After initial experiments with different variants, we choose the following models in our evaluation.
\begin{itemize}[left=1ex]
\item \textbf{resnet18\_bn (11M parameters)}: We use ResNets \cite{he2016deep} as our representative  ConvNet. In our initial experiments, larger ResNets and the other ConvNets like RegNet \cite{radosavovic2020designing} did not yield qualitative difference in performance on our datasets. We use the standard ResNet-18 architecture with one modification that we set stride of the first $7\times 7$ base convolution to $1$ rather than the standard of $2$. We do this as it gave better results with the resized datasets, specially with the group norm variation discussed below. 
\item \textbf{resnet18\_gn (11M parameters)}: It has been observed  that batch normalization often leads to poor performance in transfer learning as the batch statistics from source task could be widely off for target task (see \eg \cite{kolesnikov2020big,wu2021rethinking}). The same reasoning also applies when dealing with distribution shifts, where batch statistics could become irrelevant even when the test distribution shifts systematically \cite{djolonga2021robustness}. To overcome this, we consider a variant of \textit{resnet18}  with group normalization and weight standardization \cite{wu2018group,qiao2019weight}  in place of batch normalization. This variation indeed leads to more robust performance in our experiments, specially in the absence of data augmentation. 

\item \textbf{antialiased\_resnet18 (11M parameters)}: %
 \cite{azulay2019deep,zhang2019making} show that the downsampling layers (pooling/strides) make ConvNets non-shift invariant. To remedy this, \citet{zhang2019making} proposed a  a specialized modification to ConvNets to improve their invariance to spacial shifts by introducing a \textit{BlurPool} layer as an antialiasing-filter before downsampling. This constitute a model with more specialized priors about translation invariance built into its design. %
\item \textbf{cait\_xxs36 (17M parameters)}: CaiT architecture \cite{touvron2021going} is a variant of  the basic vision transformer (ViT) \cite{dosovitskiy2020image} that leads to more efficient training of deeper models. We use CaiT as our representative transformer model as it had the best performance in initial experiments. Other ViT variants, including larger models and the distilled variant DeiT \cite{touvron2021training} did not provide significant performance boost on our small scale datasets. %
\item \textbf{resmlp\_12 (18M parameter)}: Among the  MLP models  for image classification, we tried  MLP-mixer \cite{tolstikhin2021mlp} and ResMLP \cite{touvron2021resmlp} in our initial experiments. We stick with ResMLP for detailed experimentation as it had slightly better performance. 
\end{itemize}
Although the above model configurations are not the state-of-the-art on larger benchmarks, on smaller scale CIFAR and TinyImageNet datasets, they have competitive performance as their larger or more complex counterparts. %
Since our goal is to evaluate the relative degradation in performance with translation shifts, we do not overly optimize for top-accuracy. %

\subsection{Augmentations}
We first consider four  data augmentation pipelines while training the models described above. In the appendix, we look at more minimal augmentations to elaborate on our findings.
\begin{itemize}[left = 1ex]
\item \textbf{No Augmentation \textit{(NoAug)}}: We use this setting as a baseline for purely evaluating the merits of an architecture in generalization to translation shifts. %
\item \textbf{Basic augmentation \textit{(BA)}}: The basic augmentation consists of a random flip and a random crop with up to $4\times 4$ pixel padding. This minimal augmentation has been a de-facto standard in many vision tasks, and it already gives over $5\%$ boost in accuracy even without considering its effects on distribution shifts. Note that unlike in standard training pipeline, with our padding of training images, the random crop does not lose any original image pixels. 
\item \textbf{Advanced augmentation \textit{(AA)}}: The current slate of image data augmentation techniques are more varied and less intuitive compared to the simple transformations described above. %
In our version of advanced augmentation (AA), we use the following pipeline: first we apply (a) the basic augmentation (BA) described above, then (b) RandAugment \cite{cubuk2019autoaugment}, then (c) random erasing \cite{zhong2020random}, and finally use (d) MixUp \cite{inoue2018data}. RandAugment uses a randomly chosen composition of transformations from a predefined list. We use the standard RandAugment list from \cite{rw2019timm} but without the \texttt{TranslationX} and \texttt{TranslationY}  as  these are covered with more control within basic augmentation.
\item \textbf{AA without translation \textit{(AA(no tr))}}: %
We also consider a variant of advanced augmentation (AA) where we remove any augmentations that are explicitly related translation shifts. Specifically, we remove random crop from basic augmentation (BA) and in the RandAugment transformations list, along with previously removed \texttt{TranslationX} and \texttt{TranslationY}, we  also remove \texttt{ShearX} and \texttt{ShearY}. 
In this case, any improvements over NoAug arise only indirectly. %
\end{itemize}

\subsection{In-distribution test accuracy} 
Table~\ref{tab:in-distribution} gives the test accuracies on the in-distribution test dataset, which is preprocessed with the same padding configuration as the train dataset (\ie with $8$ pixels padded symmetrically on all sides). We use this as a reference performance without any distribution shifts (a.k.a. the \textit{in-distribution } accuracy). Ideally, we would expect a classifier that that learns good image priors to maintain their reference performance even when  the test dataset is shifted by  object invariant properties.%
\begin{table}[thb]
\centering
{\footnotesize
\begin{tabular}{|m{0.1cm}|m{1.3cm}|m{1.5cm}|m{1.5cm}|m{1.5cm}|m{1.5cm}|m{1.5cm}|} %
\toprule
&            & resent18\_bn          & resnet18\_gn      & antialiased \_resnet18    &cait\_xxs36       & resmlp\_12 \\

\midrule    
\multirow{4}{*}{\begin{turn}{270}\tiny \textbf{\!\!\!\!\!\!CIFAR-10}\end{turn}}
& NoAug       & $90.85_{\pm0.19}$    & $91.48_{\pm0.09}$  & $92.55_{\pm0.21}$    & $77.58_{\pm0.11}$ & $79.99_{\pm0.15}$ \\
& BA          & $96.10_{\pm0.05}$    & $95.96_{\pm0.06}$  & $95.63_{\pm0.15}$    & $87.69_{\pm0.50}$\remove{$88.29_{\pm0.16}$} & $87.73_{\pm0.07}$ \\
& AA(no-tr)   & $96.35_{\pm0.06}$    & $96.06_{\pm0.07}$  & $96.98_{\pm0.07}$    & $95.09_{\pm0.19}$ & $91.90_{\pm0.10}$ \\
& AA          & $97.74_{\pm0.03}$     & $98.03_{\pm0.06}$    & $97.77_{\pm0.10}$  & $97.25_{\pm0.01}$ & $96.03_{\pm0.09}$ \\
\midrule
\multirow{4}{*}{\begin{turn}{270}\tiny \textbf{\!\!\!\!\!\!CIFAR-100}\end{turn}}
& NoAug       & $67.62_{\pm0.65}$    & $64.62_{\pm0.29}$  & $70.37_{\pm0.09}$   & $43.43_{\pm0.12}$ & $52.79_{\pm0.21}$ \\
& BA          & $78.68_{\pm0.18}$    & $78.42_{\pm0.08}$  & $78.20_{\pm0.16}$   & $57.62_{\pm0.78}$ & $60.52_{\pm0.60}$ \\
& AA(no-tr)   & $74.56_{\pm0.46}$    & $74.09_{\pm0.22}$  & $77.62_{\pm0.25}$   & $77.74_{\pm1.40}$ & $65.43_{\pm0.30}$ \\
& AA          & $82.98_{\pm0.14}$    & $82.09_{\pm0.27}$  & $81.54_{\pm0.17}$   & $82.46_{\pm0.27}$ & $78.63_{\pm0.28}$ \\
\bottomrule
\end{tabular}
}
\vspace{5pt}
\caption{Accuracy on in-distribution test dataset (\ie test and train images have same padding). %
\label{tab:in-distribution}}
\end{table}

\section{Generalization to translation shifts}\label{sec:eval}
\begin{figure}[htb]
    \centering
    \includegraphics[width=\textwidth]{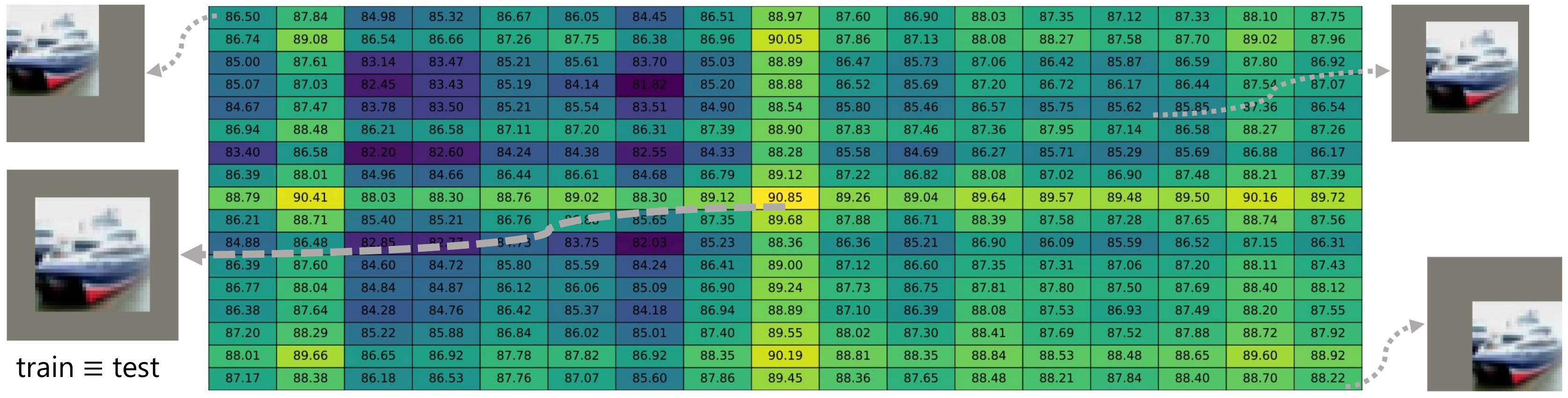}
    \caption{\textit{Generalization to translation shifts of a \textit{resnet18\_bn} trained without data augmentation (NoAug) for $1600$ epochs on CIFAR-10 dataset:} Each cell in the grid corresponds to the model performance on a test dataset with specific positioning of the object image. 
    The center cell corresponds to no translation shift from training; and the distance of the cells from the center corresponds to respective position shift between train and test dataset. The values in each cell is the accuracy on the shifted test dataset. The color corresponds to ``relative" drop in performance from the  in-distribution performance: yellow the maximum accuracy in the grid (typically the center cell), while the dark blue is saturated at $90\%$ of the max-value in the grid, \ie $10\%$ drop in accuracy. For \resbn+NoAug on CIFAR-10, max accuracy on the grid is $90.85\%$ while min accuracy is $81.82\%$.}
    \label{fig:grid-eval}
\end{figure}

With the flexibility of padded pixels in our preprocessed training data, we can now create test datasets with up to $16$ pixel translations (in Hamming distance) from the training image distribution by moving the test images anywhere within the $48\times 48$ frame. This allows us to evaluate large translation while not distorting the contents of the image itself. In the extreme locations (see, \eg corners of grid in Figure~\ref{fig:grid-eval}) there only $25\%$ overlap with the training  distribution.

For each trained model, we can look at a grid of $17\times 17$ test evaluations on modified test datasets. Each cell in the evaluation grid corresponds to the position of the $32\times 32$ test images within the $48\times 48$ frame (see illustration in Figure ~\ref{fig:grid-eval} for a \textit{resnet18\_bn} network). The center cell of the grid acts as the reference performance and corresponds to the no distribution shift, \ie the test images are centered on the frame, same as the  train images. As we move away from the center, we analogously translate the position of the object image in the test dataset. The model is then evaluated for classification accuracy on the shifted test dataset. Thus, the generalization or robustness of trained models to translation shifts can be comprehensively summarized by such a grid. 

\subsection{Case study: Convolutional networks}\label{sec:convnets}
We first describe some observations from our initial experimentation exclusively on ConvNets trained on CIFAR-10. Our evaluations on other architectures are provided in Section~\ref{sec:comparison}. In Figure~\ref{fig:grid-eval} along with the illustration of translation shift grid, we show the evaluation of \resbn  (with batch normalization) on generalization to translation shifts. This model was trained without any augmentation to demonstrate the baseline performance of convolutional architecture. 
We observe that, \emph{despite the built-in image priors in ConvNets and despite using a  weaker notion than translation invariance, there is a drop in performance on even small translations. For example, in the  non-yellow cells close to center in Figure~\ref{fig:grid-eval}, we notice that a two pixel hamming distance translation can lead to $>5\%$ drop in performance.} %
It is worth forward referencing at this point that when compared to other architectures ConvNets are relatively lot more resilient to translation shifts in the absence of data augmentation (see Section~\ref{sec:comparison})---the worst case drop in performance is $\sim10\%$ for ConvNets, while for other architectures, it could be as large as $30-50\%$ (see Figure~\ref{fig:summary}).

\begin{figure*}[thb]
    \centering
    \begin{subfigure}[b]{0.49\textwidth}
    \includegraphics[width=1.05\textwidth]{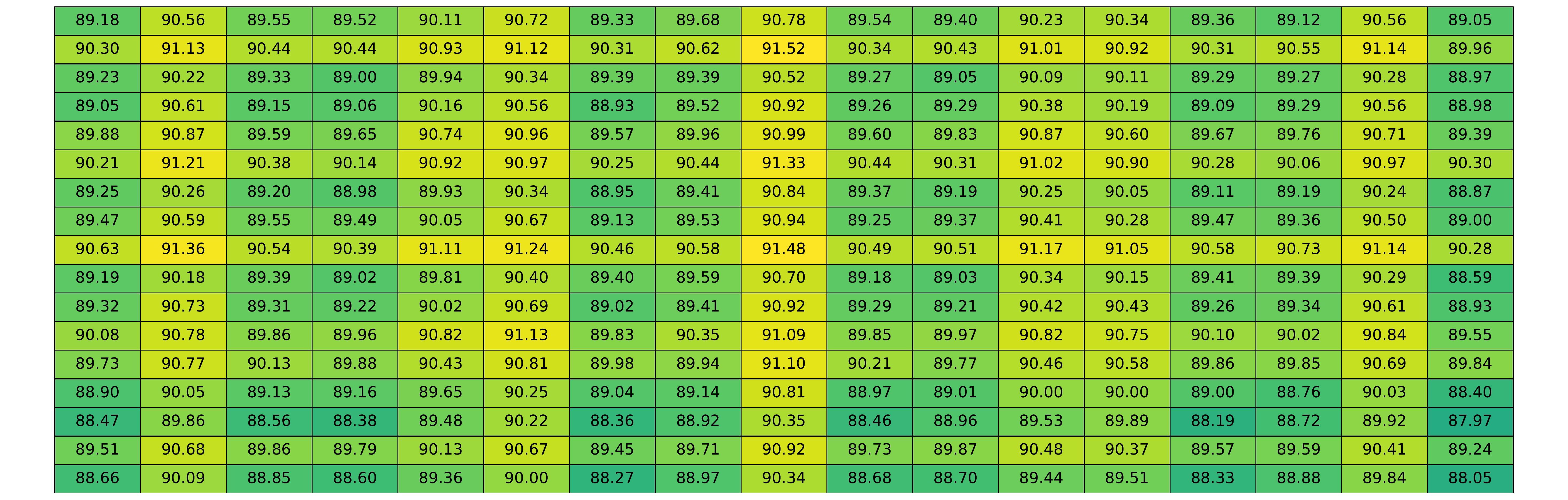}
    \caption{\centering\textit{resnet18\_gn}+NoAug: max = $91.48$, min = $87.97$. }
    \end{subfigure}
    \begin{subfigure}[b]{0.49\textwidth}
    \includegraphics[width=1.05\textwidth]{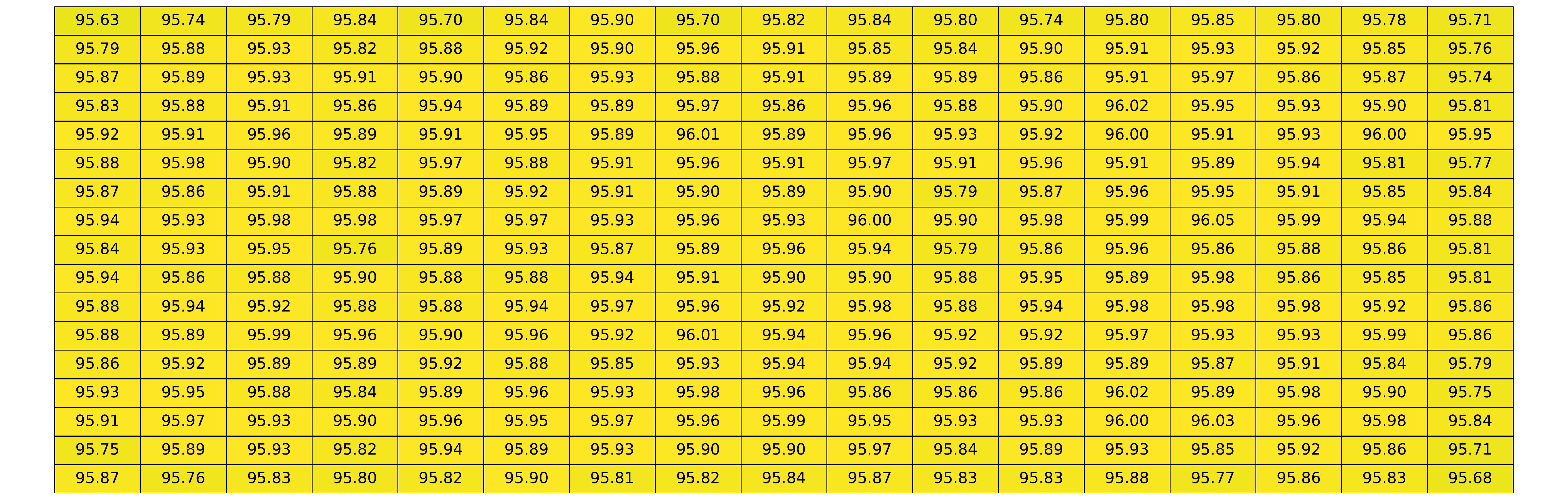}
    \caption{\centering \textit{resnet18\_gn}+BA: max = $95.96$, min = $95.63$}
    \end{subfigure}
    \caption{Generalization to translation shifts of \textit{resnet} variants using the same evaluation as in  Figure~\ref{fig:grid-eval}. (a) \textit{resnet18\_gn} network group norm and weight standardization again trained without any augmentation. (b)  \textit{resnet18\_gn} trained with basic augmentation (BA) consisting of random horizontal flip and random crop at most $4$ pixels. For quick reference, the sub-captions mention the max and min accuracy of the models over the grid. %
    \label{fig:grid-eval2}}
\end{figure*}

\paragraph{Batchnorm vs groupnorm+weight standardization.} Prior work most notably \cite{azulay2019deep,zhang2019making,chaman2021truly} attribute the lack of invariance to strides and ReLU non-linearities in the standard networks. We believe these factors also affect the weaker notion of generalization to translation shifts that we study. %
Our experiments suggests that, in addition to strides and pooling, batch normalization is yet another factor that might contribute to lack of translation invariance. Figure~\ref{fig:grid-eval2}(a) shows the performance of  \textit{resnet18\_gn} model trained with the same configuration as \resbn in Figure~\ref{fig:grid-eval}, but with a modification that all the batch normalization layers replaced by group normalization and weight standardization \cite{wu2018group,qiao2019weight}. We see that this simple modification already improves the robustness of ResNet to translation shifts. In hindsight, it is understandable that batch normalization would have detrimental effects when the test distribution shifts from the training distribution as the batch statistics obtained from exponential moving average of training statistics no longer remains accurate \cite{wu2021rethinking}. %

\paragraph{Training with basic augmentation (BA).} %
Figure~\ref{fig:grid-eval2}(b) illistrates our evaluation of \resgn with basic augmentation (BA) which includes random crops of up to $4$ pixels and horizontal flips. This simple augmentation not only improve in-distribution accuracy by over $5\%$, but also makes the models near-perfectly robust to up to $8$ pixel translations in test distribution -- indicating a form of meta-generalization from augmentations. %
In the appendix we provide further evidence of such meta-generalization: (a) In  Appendix~\ref{sec:app_convnet}, we look at even  more minimal \textit{BA-lite} and \textit{BA-liter} augmentations with smaller range of random crops of at most $2$ and $1$ pixels, respectively. There we see that even $1$ pixel augmentation improved robustness of ConvNets, while other architectures still benefit from larger random crops. (b) In Appendix~\ref{sec:tinyimagenet} our results on TinyImageNet dataset show robustness to a larger range of translation shifts of up to $16$ pixel shifts, even though BA still uses only $4$ pixel shifts.

\section{Architectures and augmentations for generalization to translation shifts}\label{sec:comparison}
\begin{figure}[h!]
    \hspace*{-1.7cm}
    \centering
    \begin{tabular}{>{\centering\arraybackslash}m{0.1\textwidth}|>{\centering\arraybackslash}m{0.46\textwidth}|>{\centering\arraybackslash}m{0.46\textwidth}|}
          & {\textbf{\large \centering CIFAR-10}} &{\textbf{\large \centering CIFAR-100}} \\
    &&\\
    \textbf{\large AA}        & \includegraphics[width=\linewidth]{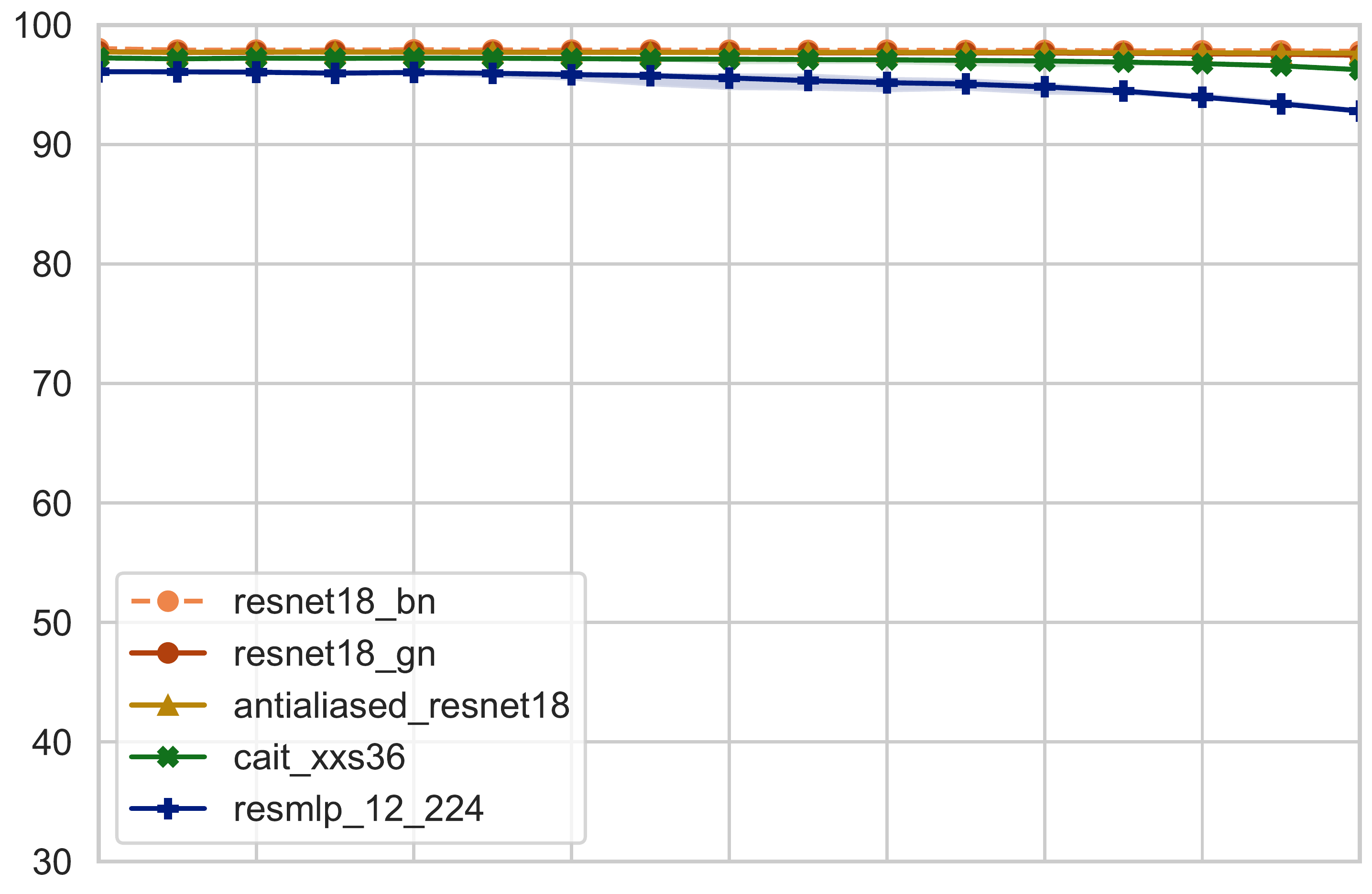}
    & \includegraphics[width=\linewidth]{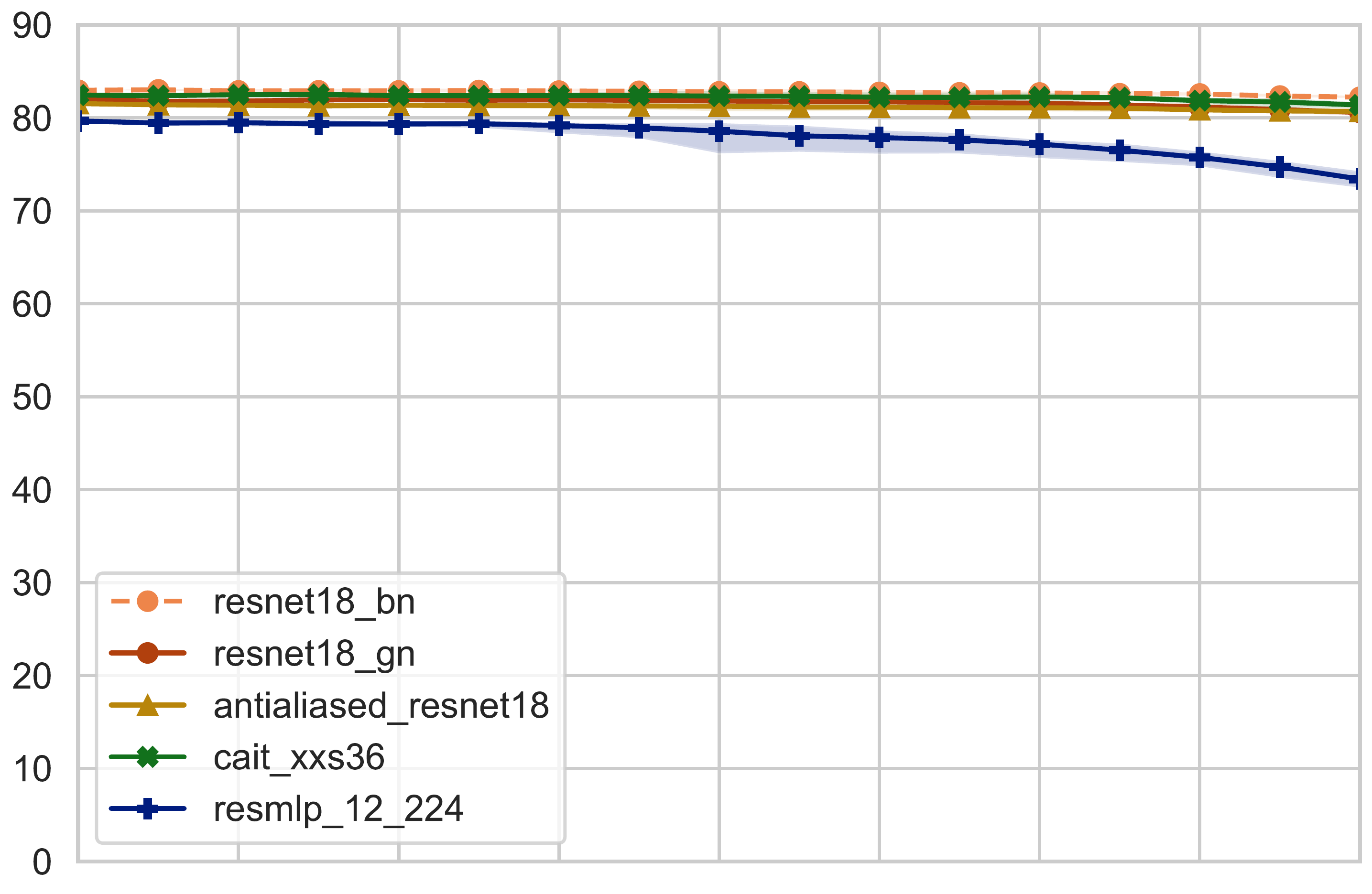}\\
    \textbf{\large BA}        & \includegraphics[width=\linewidth]{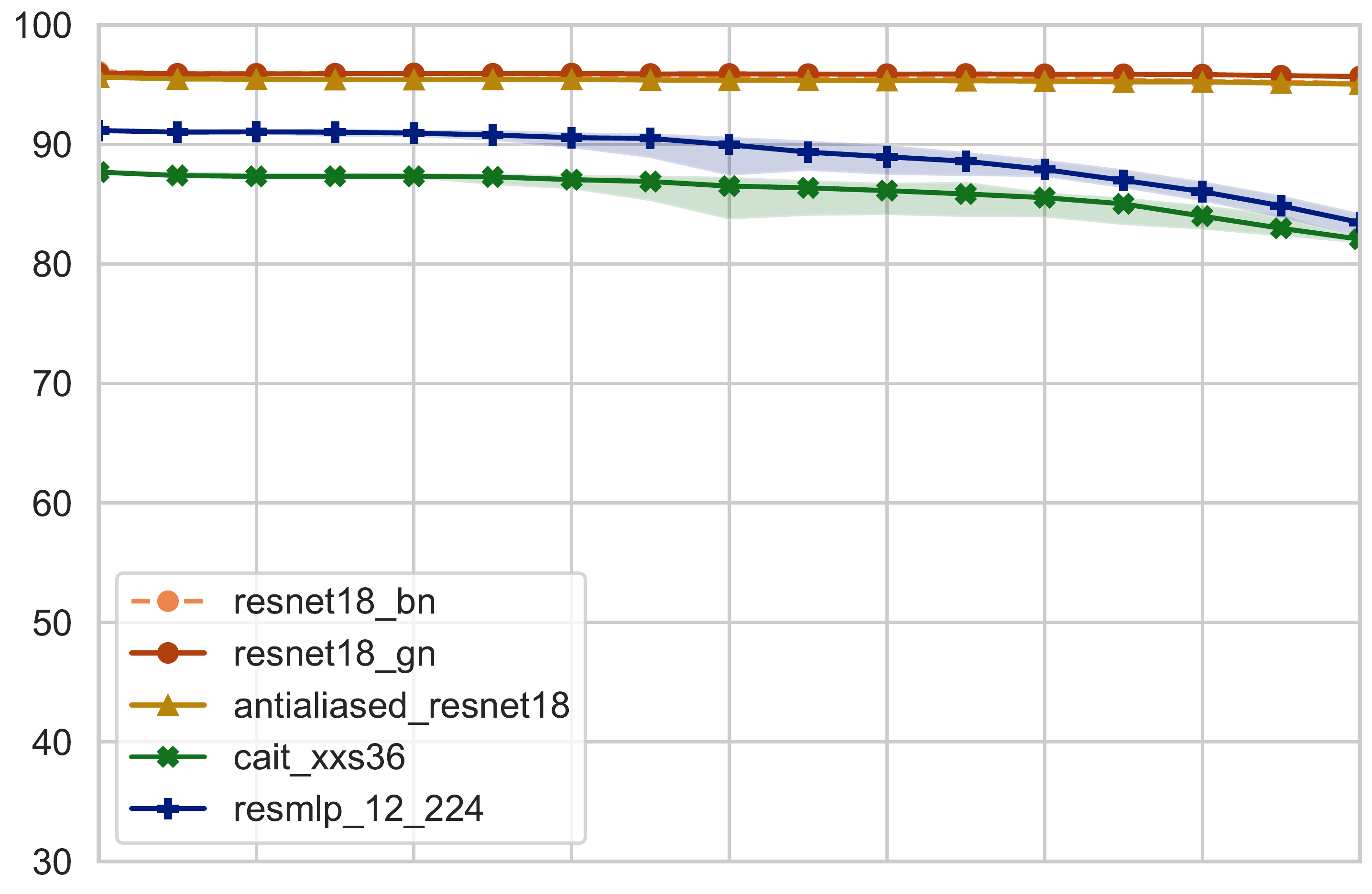}
    & \includegraphics[width=\linewidth]{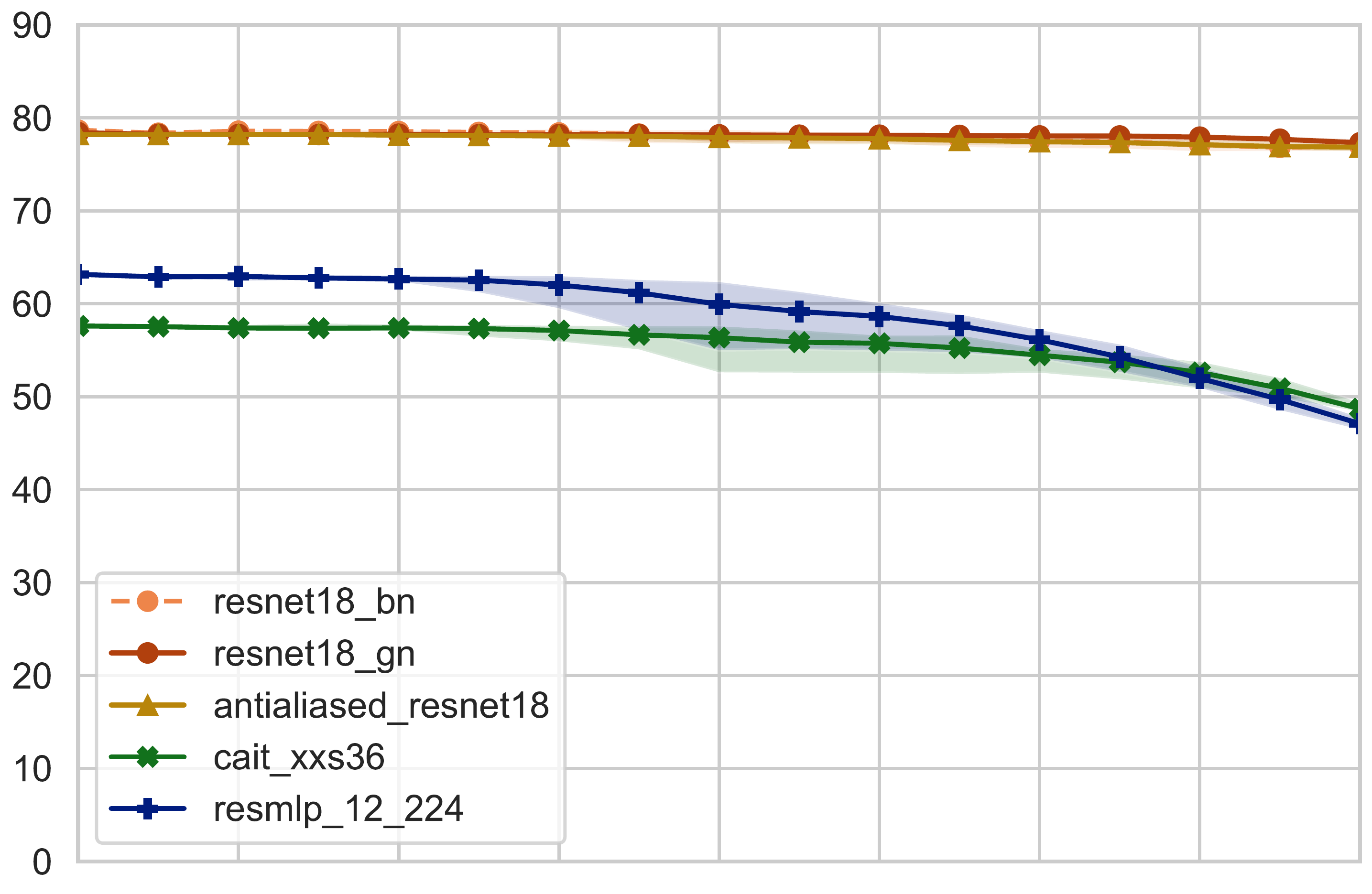}\\
    \begin{tabular}[c]{@{}c@{}}\textbf{\large AA}\\ \textbf{\large (no-tr)}\end{tabular} & \includegraphics[width=\linewidth]{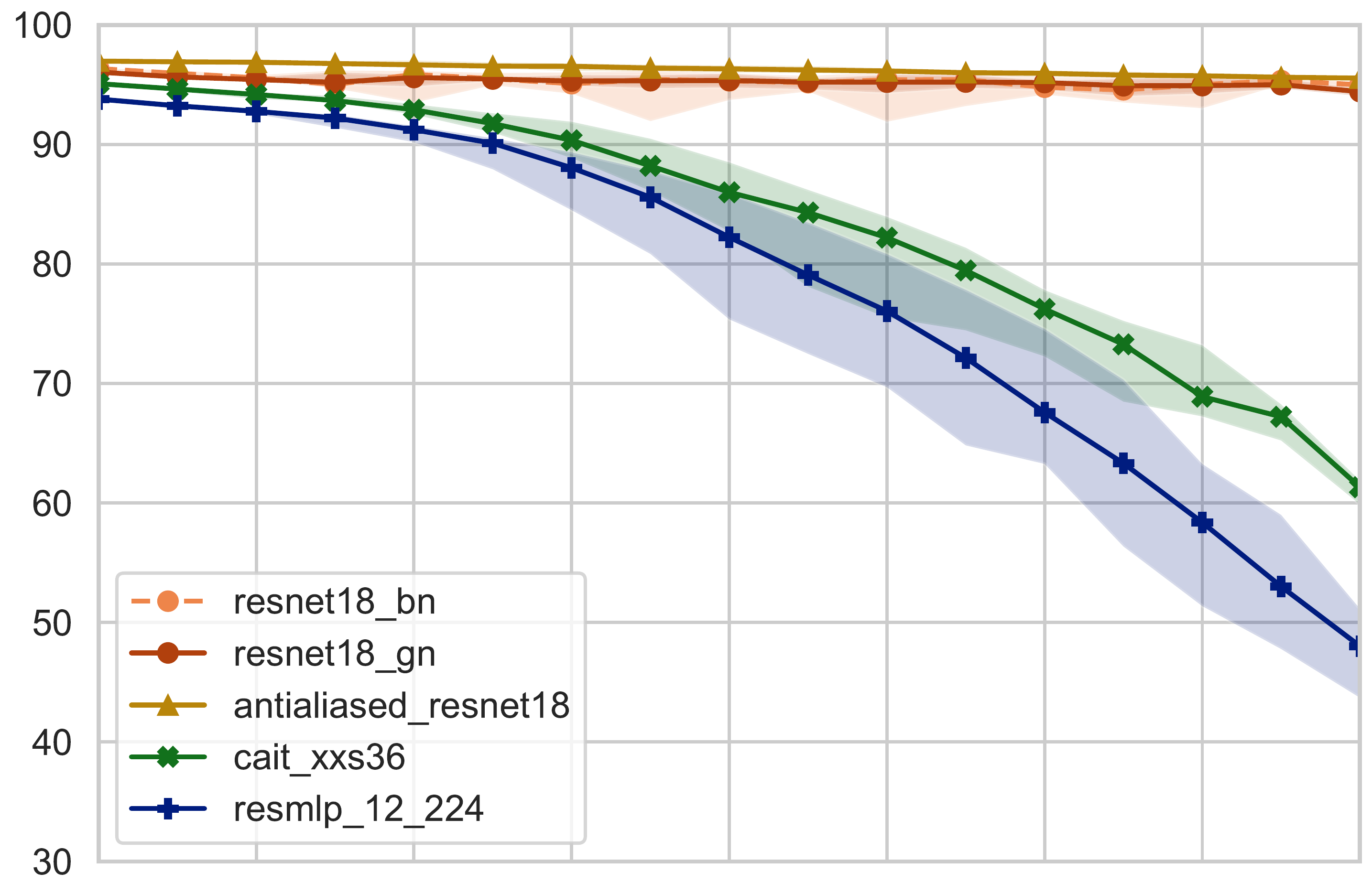}
    & \includegraphics[width=\linewidth]{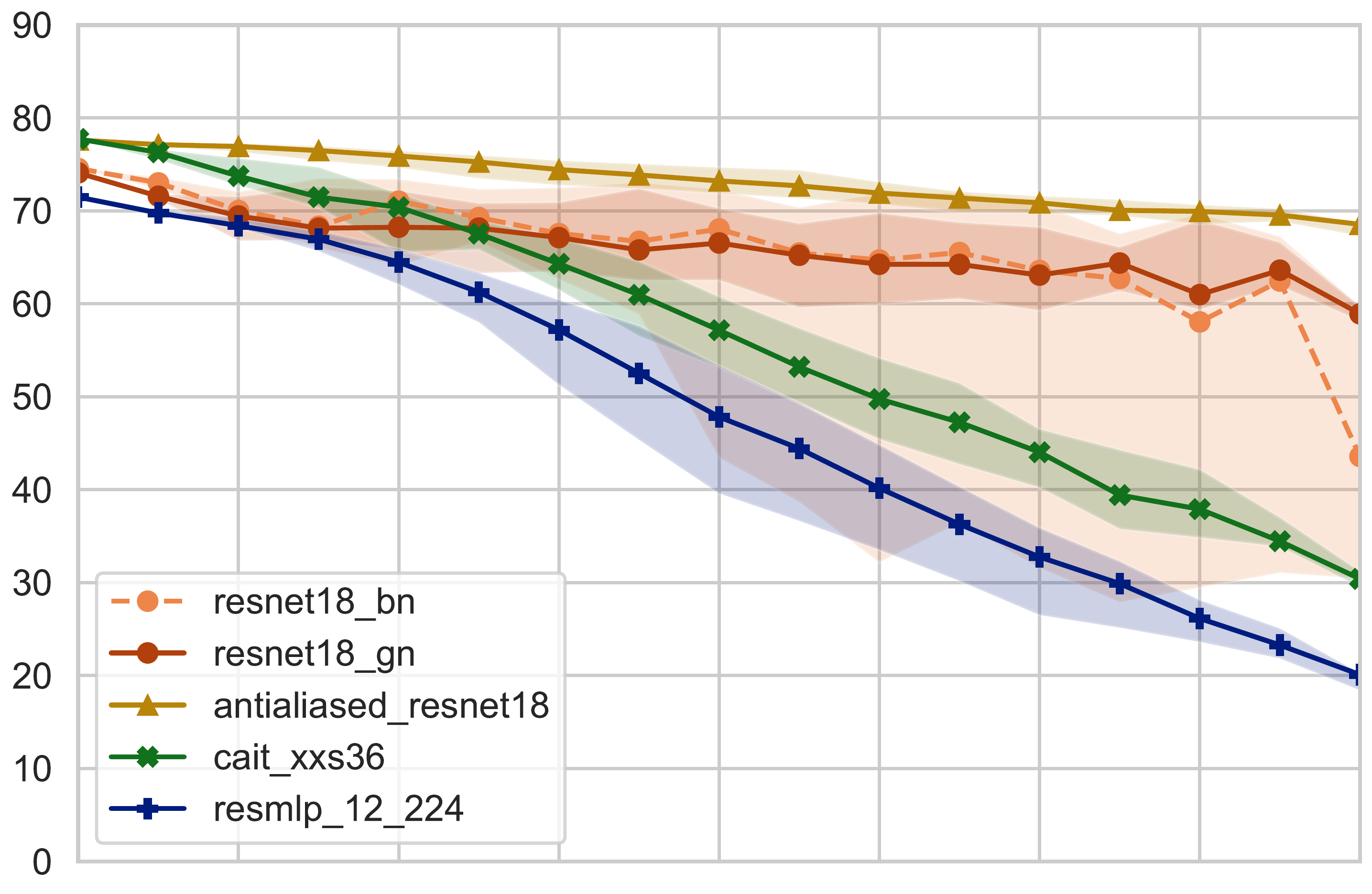}\\
    \textbf{\large NoAug}      & \includegraphics[width=\linewidth]{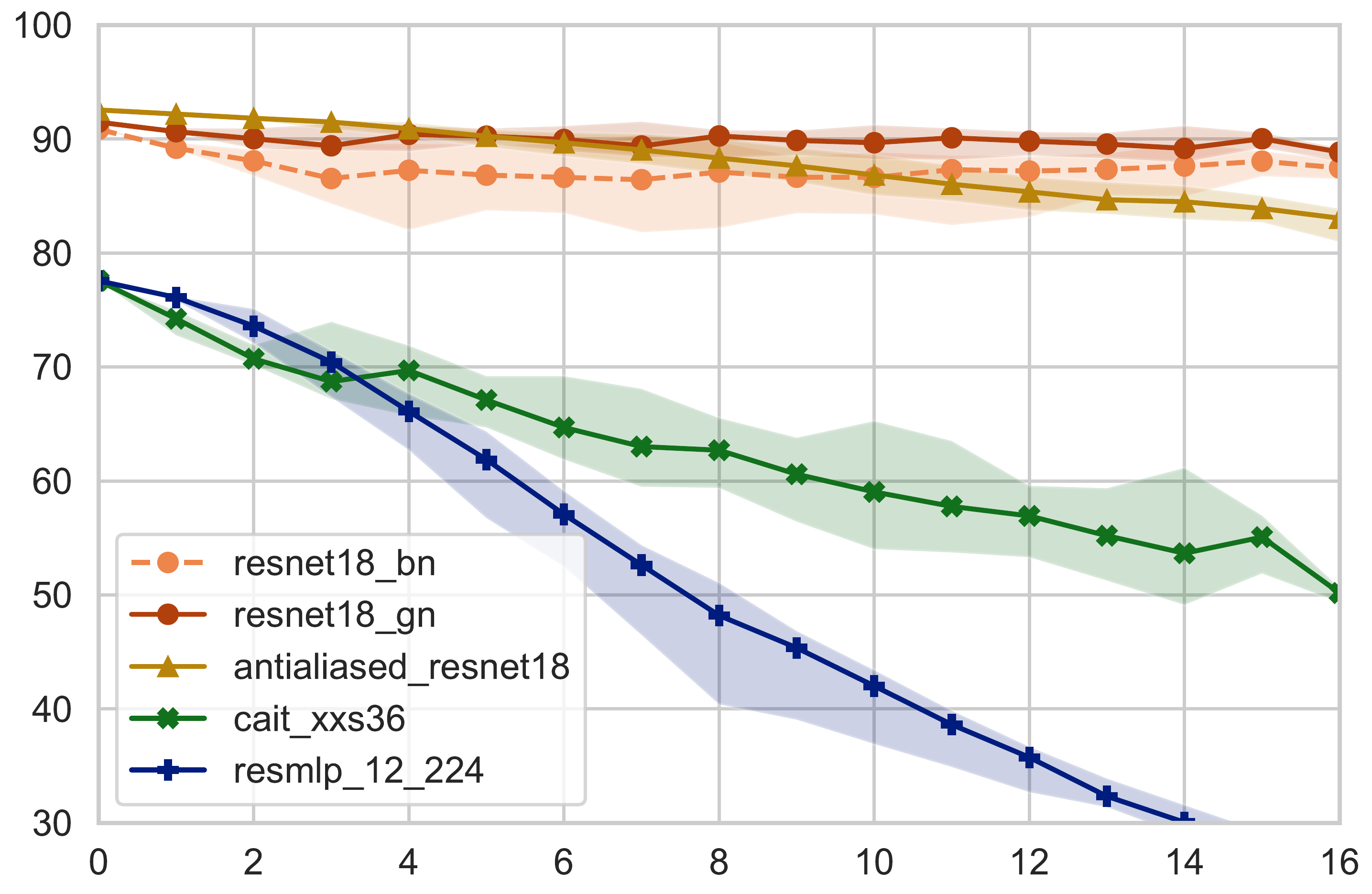}
    & \includegraphics[width=\linewidth]{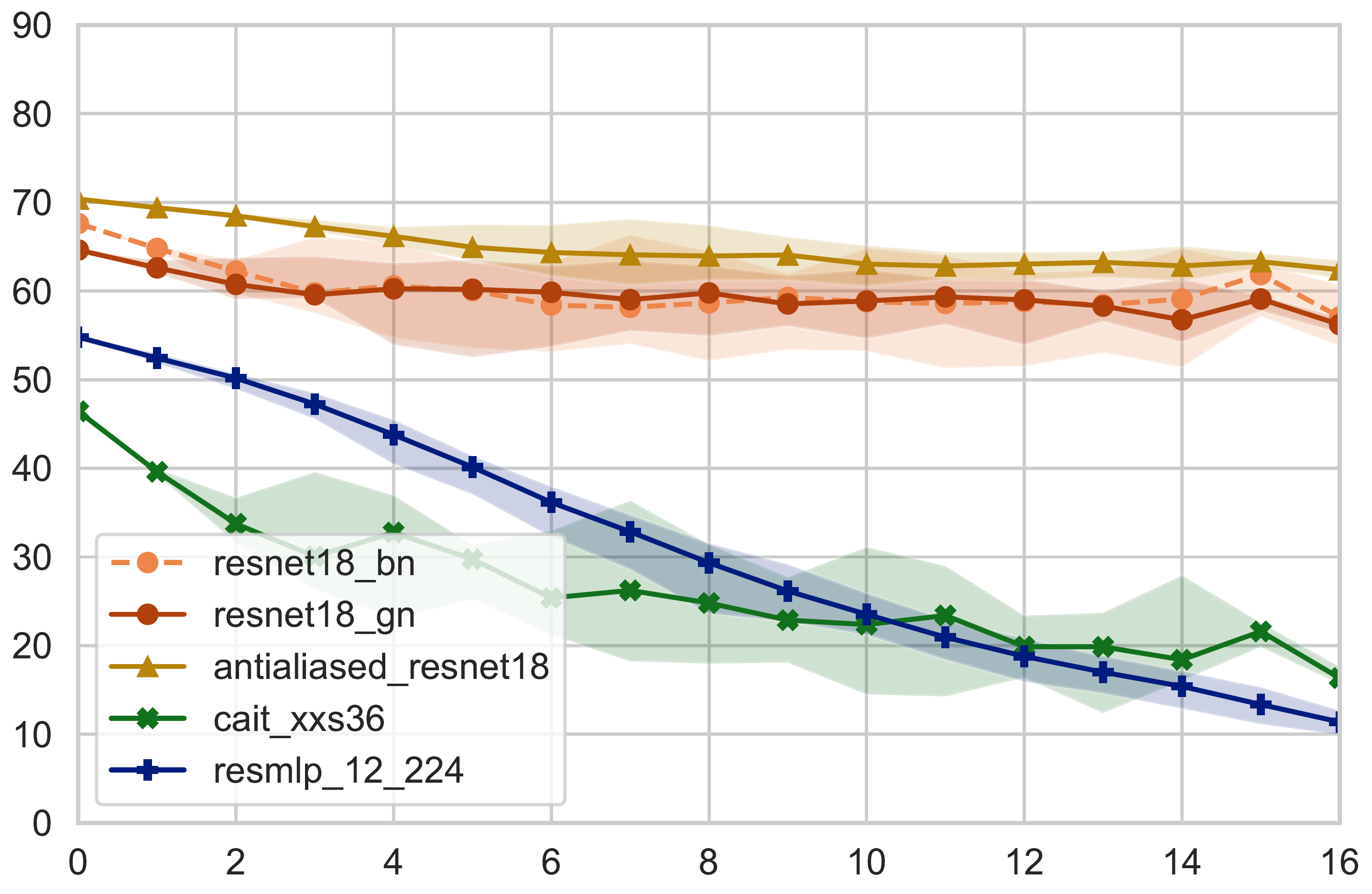}\\
    &&\\
    &&
    \end{tabular}%
    \caption{Generalization to translation shifts. The $x$-axis in each plot is the Hamming distance of the location of test images (within the $48\times 48$ frame) relative to the location in the training images (center of the frame). The larger the $x$-axis value, the larger is the shift from train distribution. For a given value of $x$, there might be multiple test configurations that are $x$-Hamming distance away (like shift of $1$ pixel to top, bottom, right, or left when $x=1$). The median of these values is plotted as line plot, while the shaded region covers the min and max values of the list. 
    The left and right plots correspond to accuracies on the CIFAR-10 and the more  challenging CIFAR-100 dataset, respectively. The $y$-axis for each dataset (column) is normalized to be on the same scale. 
    \label{fig:summary}}
\end{figure}

The summary grid view of evaluations on translation shifts (as in Figure~\ref{fig:grid-eval}-\ref{fig:grid-eval2}) is more comprehensive and we will revisit them in the appendix. However, it is not ideal for comparing different configurations of architectures and augmentations. In this section, we use an alternative visualization and plot the test accuracies as a function of Hamming distance between the position of images in the test and training datasets. The performance of all our models and augmentations are summarized in Figure~\ref{fig:summary}. %

\newpage
\subsection{Key takeaways} We make the following observations that are supported by our experiments. We provide additional experiments and discussions from our study including evaluation on TinyImageNet in the appendix. 
\begin{enumerate}[left=0ex]
\item Without data augmentation, even ConvNets with designed architecture has a noticeable drop in performance on spatially shifted test dataset. From Figures~\ref{fig:grid-eval}-\ref{fig:grid-eval2}, one can see that merely switching batchnorm with groupnorm appears to mitigate some performance gap. Using a specialized antialiased modification by \citet{zhang2019making} further improves the performance, albeit not to perfect invariance as the non-linearities could be a source of breaking spacial invariance \citet{chaman2021truly}. Nonetheless, the architectural inductive biases in these ConvNets are still useful, if not perfect, as we can predictably see that the drop in performance is much more dramatic for the non-convolutional architectures without any image priors. In CaiT and ResMLP, without augmentation, even $1-2$ pixel translation shift can lead to  dramatic drop in performance.
\item In the other extreme, with an Advanced Augmentation (AA) pipeline, all the architectures are remarkably robust even to large translation shifts in test distribution. Note that even with AA, the maximum translation augmentation we provide (in the form of random crop) is at most $4$ pixels ($8$ pixels in Hamming distance), but we see robustness to up to $8$ pixel shifts ($16$ pixels in Hamming distance). This supports a notion of meta generalization in robustness performance. In Appendix~\ref{sec:tinyimagenet}, we see that ResNets continue to be robust to even larger translation shifts of up to $16$ pixels on each direction 
on TinyImageNet dataset. To further support the idea of meta-generalization, we also show in  Appendix~\ref{sec:app_convnet}, even more minimal $1$ or  $2$ pixel random crop augmentation already boosts robustness to translation shifts. 
\item Furthermore, with AA, the performance on in-distribution test error becomes significantly closer for all architectures. Specially, the performance of resnet18 and cait\_xxs36 are  statistically identical in this setting even though we trained on the small-medium scale CIFAR datasets. The performance of resmlp\_12 is however relatively suboptimal even with AA pipeline. First, there is non-trivial gap in the in-distribution accuracies. Secondly, the robustness to translation shifts is not nearly as good as with ResNet or CaiT. Despite these differences, even for ResMLP, the augmentations dramatically boost the generalization to translation shifts and the differences in relative drop start to appear only after $10$-$12$ pixel hamming distance shifts in test distribution.  These experiments suggest that with sufficient augmentation, the relative benefits or shortcomings of the architectures are effectively diminished. 
\item Even with a minimal Basic Augmentation (BA), we  see significant improvement in robustness to translations. In fact, for ConvNets,  BA is sufficient to achieve the near perfect generalization on our canvas. This further highlights the benefit of the built-in inductive biases in ConvNet. For non-convolutional architectures, this simple augmentation is not sufficient to achieve optimal absolute test performance, but the relative robustness is still uniformly improved. 
\item Finally, an intriguing phenomenon is observed when training with advance augmentation but without translation related augmentations (AA(no-tr)). Here the absolute test accuracy of all the models improves (presumably from learning some useful priors). For ConvNets on CIFAR-10, even such indirect augmentation is effective in making the models robust to translation shifts -- but this does not appear to uniformly hold across datasets, so the conclusion might be spurious. For transformer and MLP architectures, the robustness does not improve significantly, even though the in-distribution accuracies are significantly higher.
\item Somewhat tangentially, our experiment slightly supports the position that data augmentation can recover some of the benefits of large datasets even when learning with general architectures like ViT and MLP are beneficial. In ImageNet scale datasets this was previously observed in \citet{touvron2021going,steiner2021train} for ViTs and \citet{touvron2021resmlp} for MLPs. Our experiments show similar validation even on small CIFAR-10 datasets.%
\end{enumerate}

In summary, even though convolutional networks are not invariant or robust to translations in absolute sense, they clearly fare much better compared to other general-purpose architectures. Specially, ConvNets learn robust models with minimal augmentation, while it appears the transformer and MLP architectures require more sophisticated augmentations. At the same time, our experiments suggest that data augmentation can enforce learning of the right inductive bias with comparable or more effectiveness than the network architecture. %

\clearpage

\bibliographystyle{plainnat}
\bibliography{ref.bib}
\clearpage

\appendix

\section{Additional experimental setup, training, and hyperparameter details}
CIFAR-10 and CIFAR-100 datasets used in the main paper are standard small-medium scale benchmarks for image classification consisting of 50K training images and 10K test images evenly distributed across $10$ and $100$ classes, respectively. We recall from the main paper that the CIFAR images of size  $32\times 32\times 3$ are preprocessed using a mean padding canvas of $8$ pixels on each size (leading to $48\times 48\times 3$ images),  and  resizing to standard ImageNet training size of $224\times224\times 3$ using bilinear interpolation (see Figure~\ref{fig:samples}). 

In addition to the experiments on CIFAR datasets, we also extend our study to TinyImageNet \citep{le2015tiny} dataset in the Section~\ref{sec:tinyimagenet}. TinyImageNet is a smaller subset of the standard ImageNet dataset consisting of 100K training images and 10K validation images evenly distributed among 200 classes. Further, the  images are resized to $64\times64\times 3$ from the standard ImageNet size. This dataset curated for research exploration, with the goal of having the complexity of ImageNet objects, while keeping the computation manageable. However, since the size of TinyImageNet is an order of magnitude smaller than ImageNet, the state-of-the-art in-distribution test accuracies on this dataset are not as good as those on ImageNet. On the other hand, the images have more diverse than CIFAR datasets. Thus, this dataset is a good intermediate testbed where the computation is reasonable at the same time the objects are more diverse are not focused tightly in the center as is usual in CIFAR. 

We pass the TinyImageNet dataset through a similar preprocessing pipeline as CIFAR dataset, except now we use an even larger padding canvas of $16$ pixels ($1/4$ of image size) on each side leading to padded images of size $96\times 96\times 3$ -- the larger padding maintains the ratio of size of translation shifts we evaluate to $1/4$ of image dimension in each direction as in CIFAR dataset. Once again to avoid extensive hyperparameter tuning on transformer and MLP models, we resize the padded images  to $224\times224\times 3$ using bilinear interpolation.

The optimization algorithms and hyperparameter used for training the  models in the main paper are summarized in Table~\ref{table:hyperparameters}.  Additionally, we use the following  advanced augmentation parameters for models trained with AA and AA(no tr), where all the implementations are from \cite{rw2019timm}: For RandomAugment we use  $M=5, N=2$. Random erasing is used with probability $=0.25$. Mixup is used in batch mode with probability $1$, where switch between standard mixup and cutmix with probability $0.5$ each time; mixup $\alpha=0.8$ and mixup label smoothing $=0.1$. We also use the repeated augmentation \cite{hoffer2020augment} when training with AA or AA(no tr) in distributed mode.

\begin{table}[thb]
\centering
    \begin{tabular}{|l|c|c|c|}
        \toprule
                    & resnet18(\_bn,\_gn)   & cait\_xxs36   & resmlp\_12\\
        \midrule
        opt. alg.   & SGD+momentum          & AdamW         & LAMB\\
        batchsize   & $128\times 8$         & $64\times 8$   & $128\times 8$\\
        starting lr & $0.1\times\frac{\text{batchsize}}{512}$ & $0.001\times\frac{\text{batchsize}}{512}$ & $0.005\times\frac{\text{batchsize}}{512}$ \\
        weight decay & $5e^{-4}$            & $0.05$        & $0.05$\\
        drop path   & none & 0.1 &0.1\\
        dropouy     & 0 & 0 & 0 \\
        gradient clip & yes                 & no            & no\\
        \bottomrule
    \end{tabular}
    \vspace{5pt}
    \caption{Summary of training algorithm and hyperparameters. All models are trained distributed on $8\times$V100 GPUs and we use the automatic mixed precision  implementation through the torch.cuda.amp library in pytorch  for ResNet and CaiT training. We train all models for $1600$ epochs (except for one experiment in Section~\ref{sec:app_convnet}) and we use cosine annealing with $T_{\text{max}}=\#epochs$ and warmup for $20$ epochs to the linearly scaled starting learning rate. When using gradient clipping, we clip to norm $2$ for CIFAR datasets and $5$ for TinyImageNet. %
    \label{table:hyperparameters}}
\end{table}

\section{Additional experiments exploring meta generalization}\label{sec:app_convnet}
In the main paper, we observed that even though all our augmentation pipelines are restricted to at most $4$ pixel random crops, the learned models display robustness to twice larger translation shifts of up to $8$ pixels. We termed this phenomenon as meta generalization, where in small magnitude augmentations of an invariance property leads to generalization to much larger shifts in the test distribution. To investigate this phenomenon more, we introduce $4$ additional augmentation protocols that severely restrict the amount of direct augmentation we get about translation invariance. 

\begin{figure}[htb]
    \hspace*{-1.7cm}
    \centering
    \begin{tabular}{>{\centering\arraybackslash}m{0.1\textwidth}|>{\centering\arraybackslash}m{0.46\textwidth}|>{\centering\arraybackslash}m{0.46\textwidth}|}
          & {\textbf{\centering CIFAR-10}} &{\textbf{\centering CIFAR-100}} \\
    &&\\
    \textbf{\large BA}   
    & \includegraphics[width=0.9\linewidth]{figs/deltasum_absolute_translation_DA.pdf}
    & \includegraphics[width=0.9\linewidth]{figs/cifar100_deltasum_absolute_translation_DA.pdf}\\
    \begin{tabular}[c]{@{}c@{}}\textbf{\large BA}\\ \textbf{\large lite}\end{tabular}
    & \includegraphics[width=0.9\linewidth]{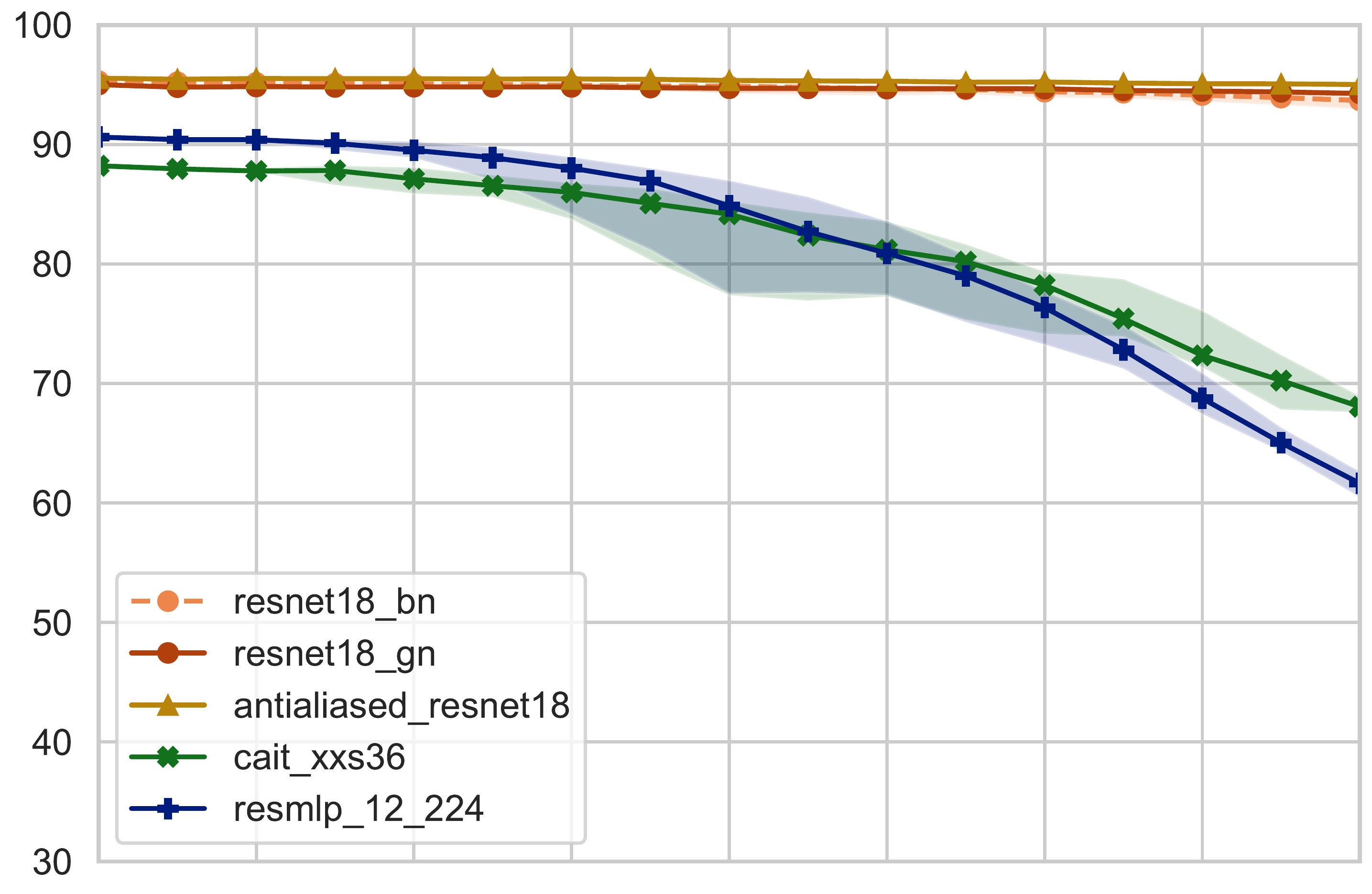}
    & \includegraphics[width=0.9\linewidth]{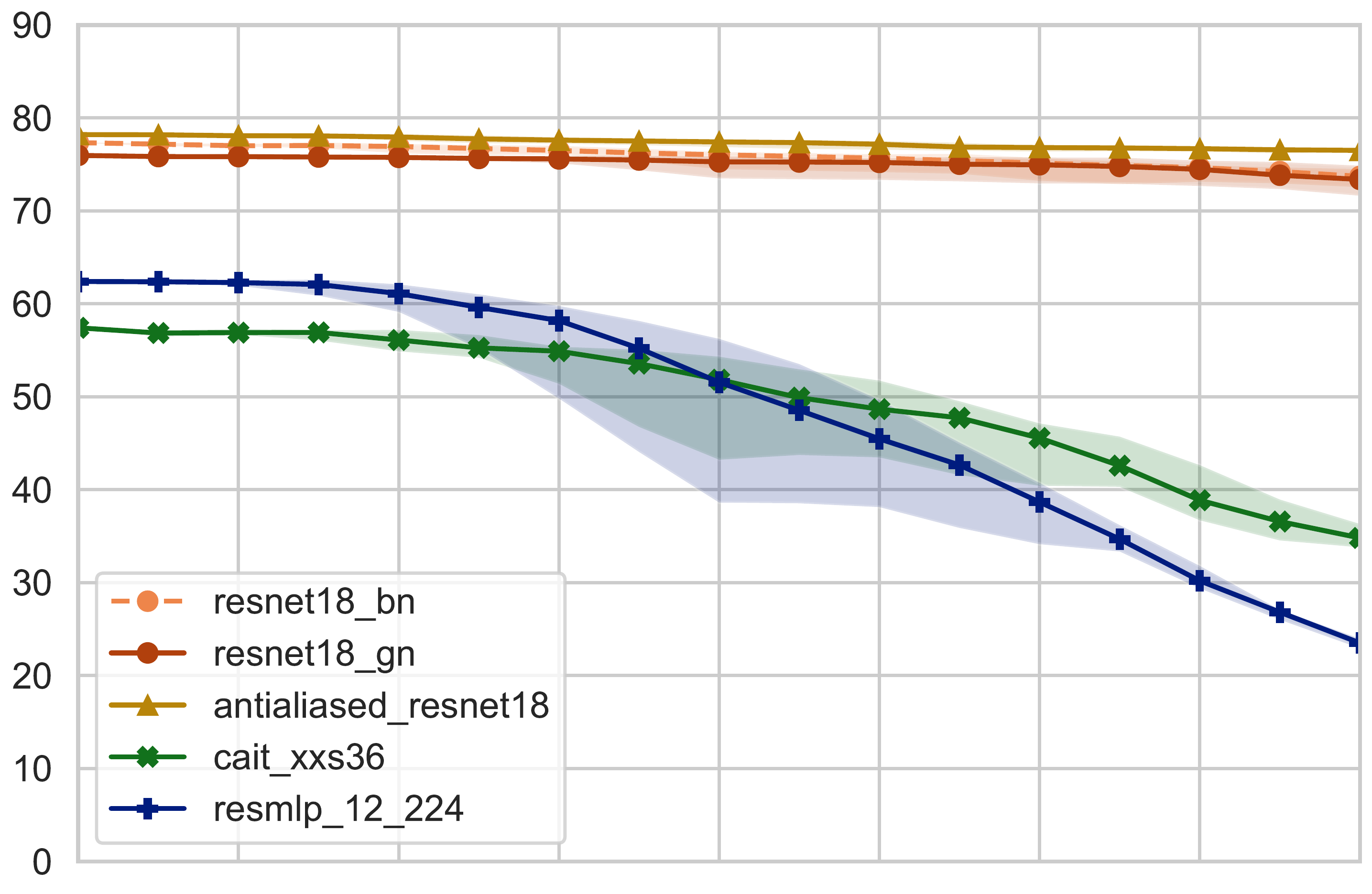}\\
    \begin{tabular}[c]{@{}c@{}}\textbf{\large BA}\\ \textbf{\large liter}\end{tabular}
    & \includegraphics[width=0.9\linewidth]{figs/deltasum_absolute_translation_DA2.pdf}
    & \includegraphics[width=0.9\linewidth]{figs/cifar100_deltasum_absolute_translation_DA2.pdf}\\
    \textbf{\large NoAug}      
    & \includegraphics[width=0.9\linewidth]{figs/deltasum_absolute_translation.pdf}
    & \includegraphics[width=0.9\linewidth]{figs/cifar100_deltasum_absolute_translation.pdf}\\
    &&\\
    &&
    \end{tabular}%
    \caption{Minimal augmentation pipelines: Basic Augmentation (BA) variants. The figure is plotted with the same protocol as the summary in Figure~\ref{fig:summary}. BA-lite and BA-liter correspond to AA pipeline with at most $2$ pixel and $1$ pixel random crop respectively.\label{fig:balite}}
\end{figure}

\begin{figure}[htb]
    \hspace*{-1.7cm}
    \centering
    \begin{tabular}{>{\centering\arraybackslash}m{0.1\textwidth}|>{\centering\arraybackslash}m{0.46\textwidth}|>{\centering\arraybackslash}m{0.46\textwidth}|}
          & {\textbf{\small \centering CIFAR-10}} &{\textbf{\small \centering CIFAR-100}} \\
    &&\\
    \textbf{\large AA}        
    & \includegraphics[width=0.8\linewidth]{figs/deltasum_absolute_translation_AA.pdf}
    & \includegraphics[width=0.8\linewidth]{figs/cifar100_deltasum_absolute_translation_AA.pdf}\\
    \begin{tabular}[c]{@{}c@{}}\textbf{\large AA}\\ \textbf{\large lite}\end{tabular}
    & \includegraphics[width=0.8\linewidth]{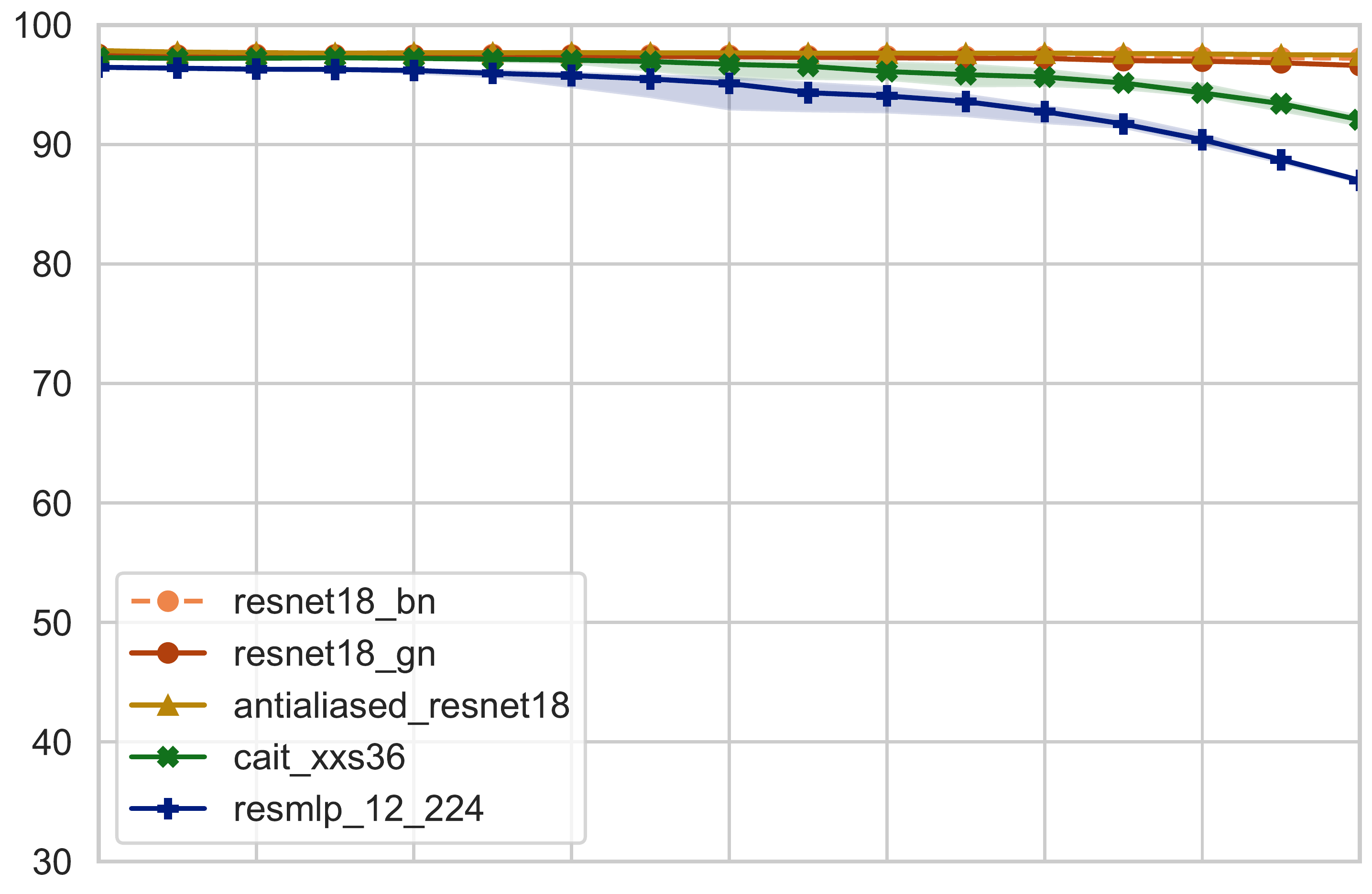}
    & \includegraphics[width=0.8\linewidth]{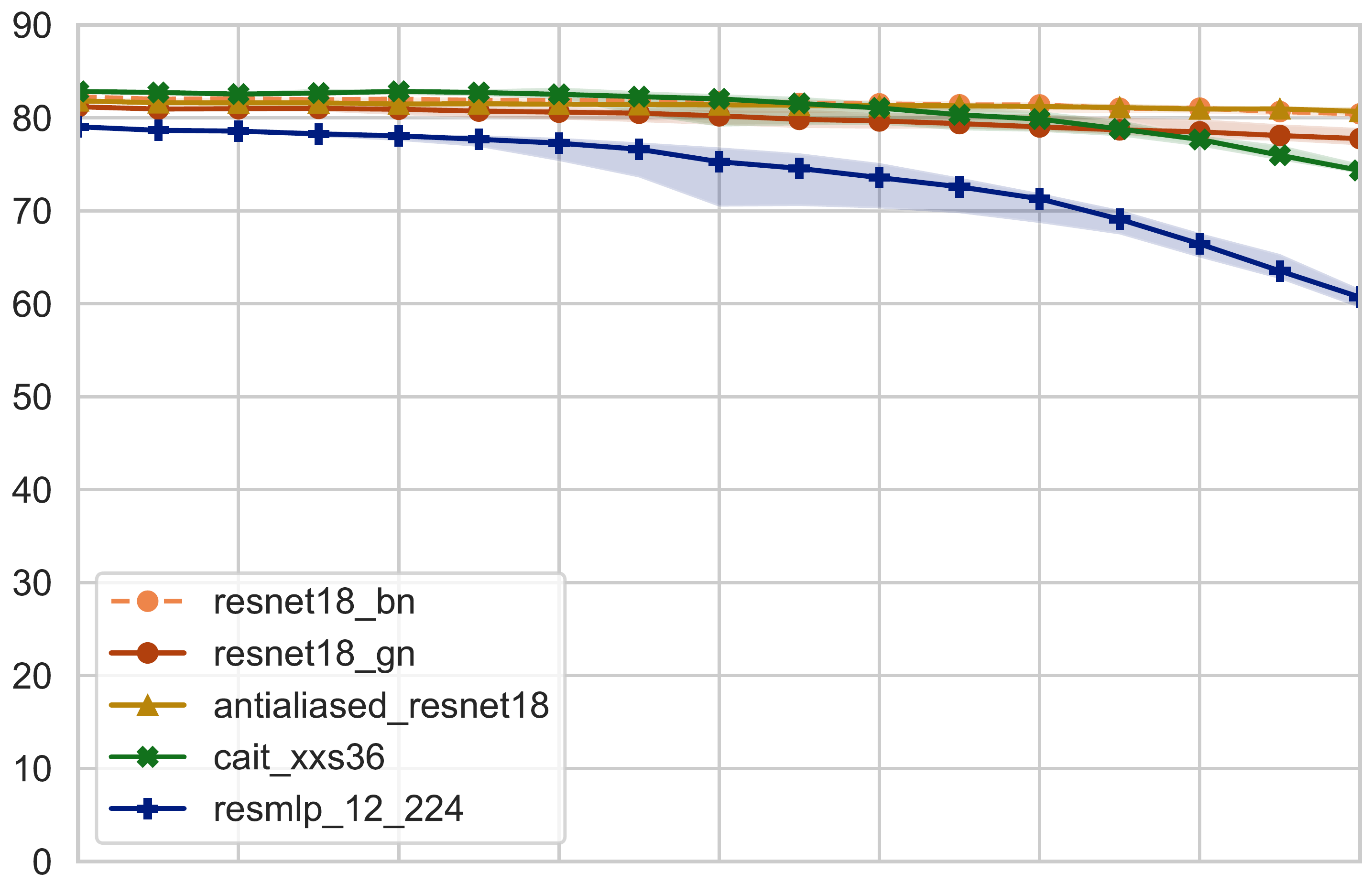}\\
    \begin{tabular}[c]{@{}c@{}}\textbf{\large AA}\\ \textbf{\large liter}\end{tabular}
    & \includegraphics[width=0.8\linewidth]{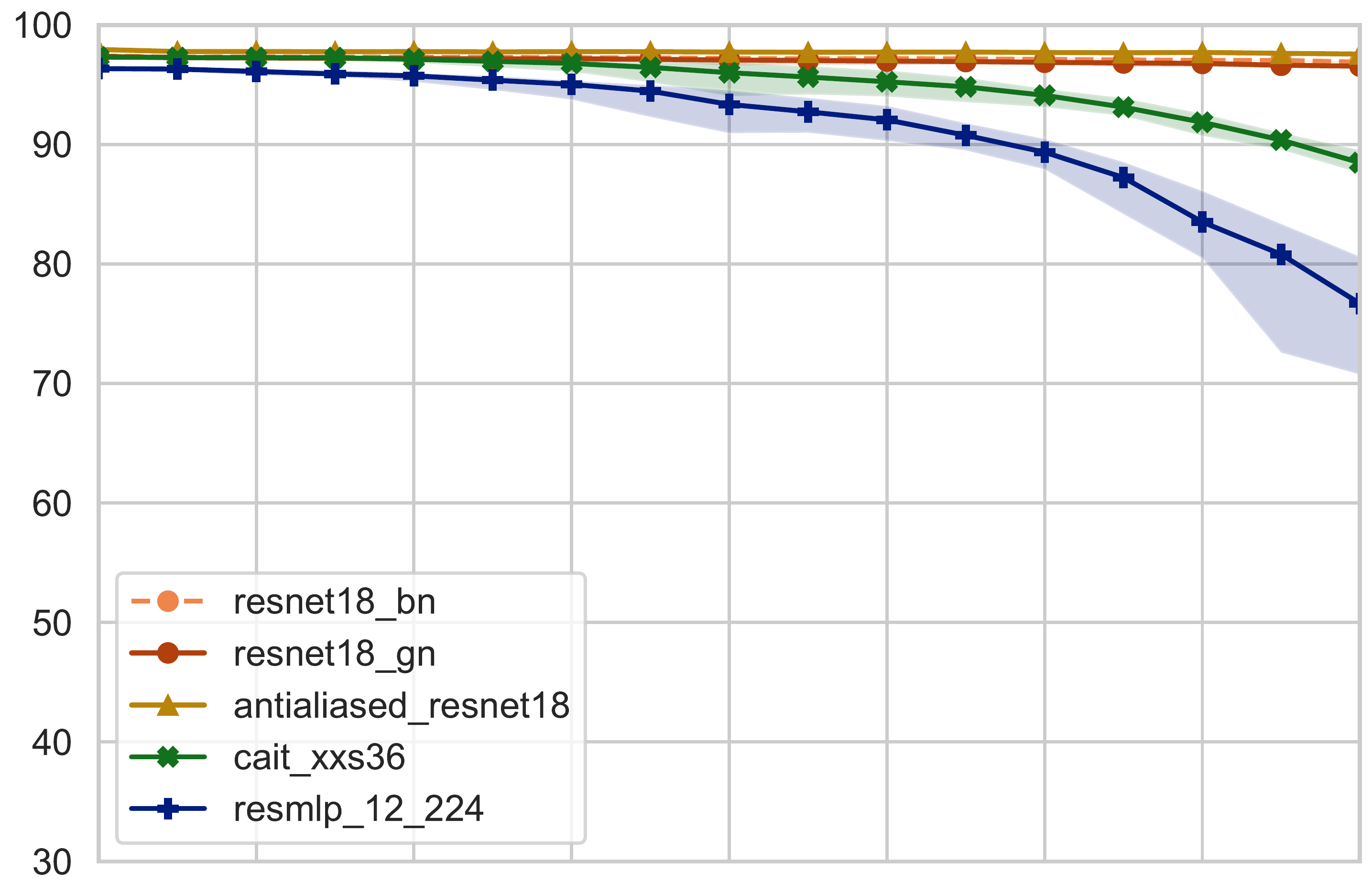}
    & \includegraphics[width=0.8\linewidth]{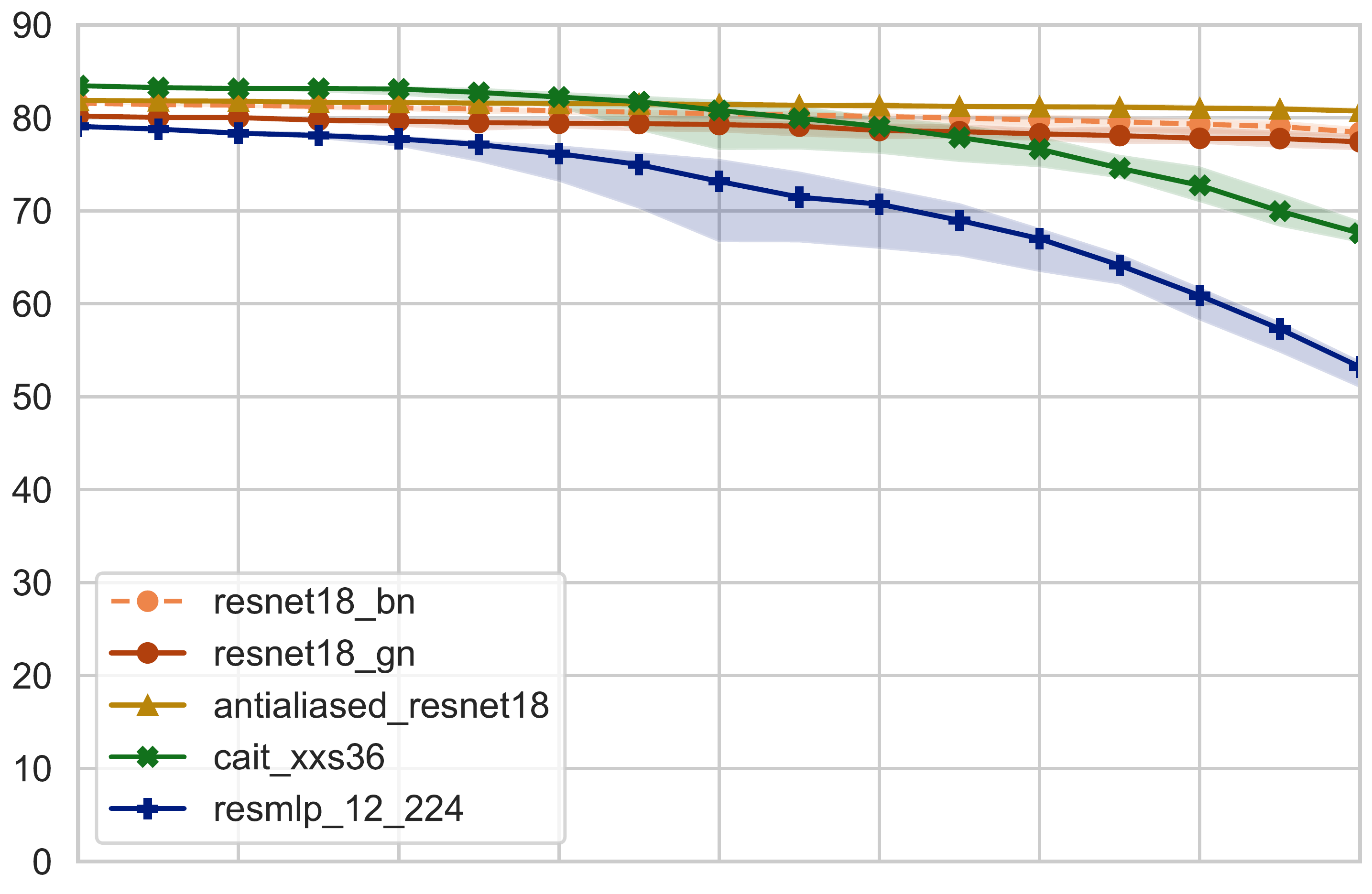}\\
    \textbf{\large NoAug}      
    & \includegraphics[width=0.8\linewidth]{figs/deltasum_absolute_translation.pdf}
    & \includegraphics[width=0.8\linewidth]{figs/cifar100_deltasum_absolute_translation.pdf}\\
    &&\\
    &&
    \end{tabular}%
    \caption{Minimal augmentation pipelines: Advanced Augmentation (AA) variants. The figure is plotted with the same protocol as the summary in Figure~\ref{fig:summary}. AA-lite and AA-liter correspond to AA pipeline with at most $2$ pixel and $1$ pixel random crop respectively.\label{fig:aalite}}
\end{figure}

\begin{enumerate}
    \item \textbf{BA-lite}  We use  basic augmentation (BA) but with random crop of only up to $2$ pixels on each side along with random horizontal flip.
    \item \textbf{BA-liter}  A more extreme BA where we restrict the random crop to only $1$ pixel pad. %
    \item \textbf{AA-lite} Here we keep the entire advance augmentation (AA) pipeline but analogous to BA-lite, restrict the random crop to at most $2$ pixel padding.
    \item \textbf{AA-liter} This  is the same as AA-lite but with at most $1$ pixel random crop augmentation. 
\end{enumerate}
We summarize the results of the above mentioned minimal augmentation protocols in Figure~\ref{fig:balite} and Figure~\ref{fig:aalite}, respectively, for the BA and AA pipelines.

In  Figure~\ref{fig:balite} we see that ConvNets when trained with even an extremely minimal augmentation of BA-liter (at most $1$ pixel random crop) show remarkably performance gains. They not only achieve near-perfect generalization to translation shifts of up to $8$ pixels, but also increase their absolute accuracy and recover most of the  performance gains  from standard basic augmentation (BA), wwhich uses up to $4$ pixel random crop. 

Even for non-convolutional architectures, the gains from minimal basic augmentations (BA-lite and BA-liter) compared to no augmentation are substantial. More impressive gains are obtained from AA-lite and AA-liter pipelines. However, unlike ConvNets the robustness is not perfect and these architectures continue to benefit from more aggressive augmentation.  We do emphasize that there is still evidence of meta generalization in these non-convolutional models: recall that even in the full advanced augmentation pipeline (AA) has at most $4$ random crop. In the following section we further see general purpose models, specially CaiT, trained with AA continues to be robust even larger shifts of up to $16$ pixels on TinyImagenet dataset.

\remove{
resnet18_BN: median(Prec@1) bestepoch: min=81.82, (0,0)=90.85, max=90.85
resnet18_DA_BN: median(Prec@1) bestepoch: min=94.87, (0,0)=96.1, max=96.1
resnet18_AAtr0_BN: median(Prec@1) bestepoch: min=91.91, (0,0)=96.35, max=96.35
resnet18_AA_BN: median(Prec@1) bestepoch: min=97.71, (0,0)=98.03, max=98.03
resnet18_DA2_BN: median(Prec@1) bestepoch: min=94.46, (0,0)=96.01, max=96.03
resnet18_BN_400: median(Prec@1) bestepoch: min=84.35, (0,0)=90.49, max=90.49

resnet18_GN: median(Prec@1) bestepoch: min=87.97, (0,0)=91.48, max=91.52
resnet18_DA_GN: median(Prec@1) bestepoch: min=95.63, (0,0)=95.96, max=96.05
resnet18_AAtr0_GN: median(Prec@1) bestepoch: min=94.06, (0,0)=96.06, max=96.06
resnet18_AA_GN: median(Prec@1) bestepoch: min=97.34, (0,0)=97.77, max=97.86
resnet18_DA2_GN: median(Prec@1) bestepoch: min=95.71, (0,0)=96.12, max=96.12
resnet18_GN_400: median(Prec@1) bestepoch: min=82.79, (0,0)=91.15, max=91.15

}

\section{Generalization to translation shifts: experiments on TinyImageNet}\label{sec:tinyimagenet}
\begin{figure}[htb]
\hspace{-1.3cm}
\centering
    \begin{tabular}{>{\centering\arraybackslash}m{0.07\textwidth}>{\centering\arraybackslash}m{0.92\textwidth}}
    \textbf{\large AA}        &
    \includegraphics[width=\linewidth]{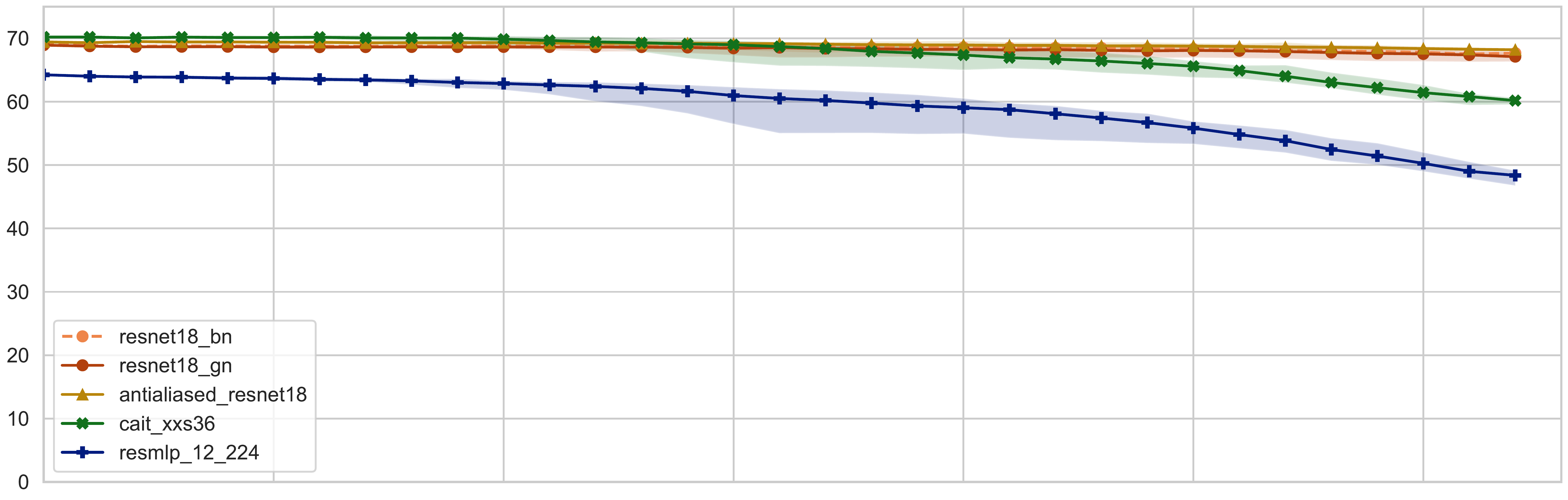}\\
    \textbf{\large BA}        &
    \includegraphics[width=\linewidth]{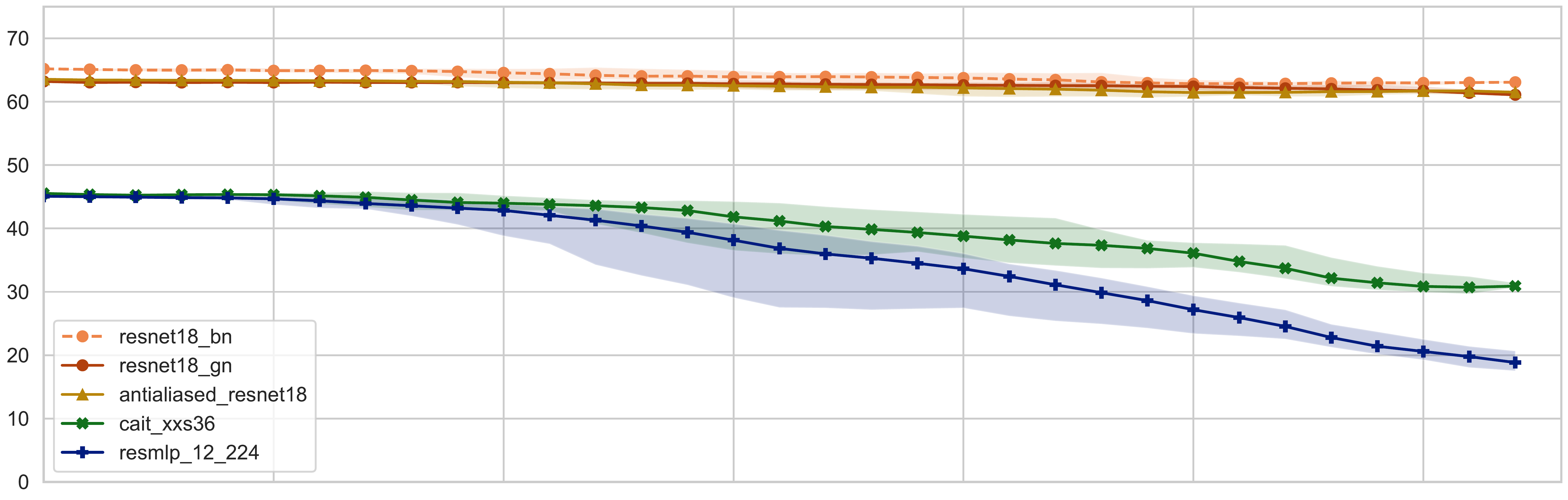}\\
    \textbf{\large NoAug}      & 
    \includegraphics[width=\linewidth]{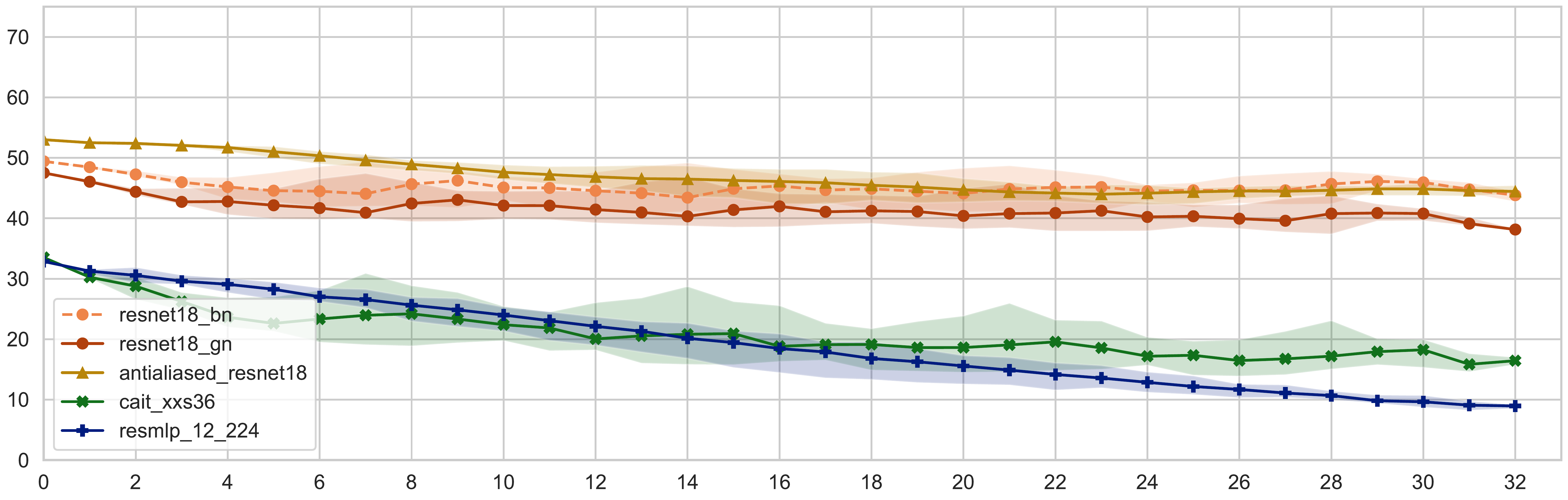}\\
    \end{tabular}%
    \caption{Generalization to translation shifts results on TinyImageNet.  The plots are of the same format as Figure~\ref{fig:summary}. The $x$-axis is the Hamming distance between the position of test images (within the canvas of $16$ pixels on each side) and corresponding position of the training images. For a given value of $x$, there might be multiple test configurations that are $x$-Hamming distance away, the median of these values is plotted as line plot, while the shaded region covers the min and max values of the list. 
\label{fig:tinyimagenet eval}}
\end{figure}

In this section, we repeat the experiments in Section~\ref{sec:comparison} on TinyImageNet dataset. In comparison to CIFAR datasets, TinyImageNet more diverse images where the objects are often not as centered. A priori, thus we expect less sensitivity to small translations and shifts. Another difference from our CIFAR evaluation is that in order to maintain the relative ratio of padding to original image, we pad the $64\times64$ images with a larger canvas of $16$ pixels rather than the $8$ pixels used in CIFAR datasets (both are padded with $1/4$th of image size). This allows us to study the effects of even larger translation shifts. Figure~\ref{fig:tinyimagenet eval} shows results on TinyImageNet that are analogous to Figure~\ref{fig:summary} on CIFAR-10 and CIFAR-100, except that we drop AA (no tr) augmentation as it did not yield meaningful insights even in CIFAR-10.

Many of our observations on CIFAR also carry over to TinyImageNet. We highlight some key similarities and differences below. In addition, we notice that in comparison to CIFAR datasets, the drop in performance appears to have a smaller slope w.r.t the pixel shifts in hamming distance. We attribute this to the TinyImageNet dataset having more diversity in the positions of objects. 

\begin{enumerate}
    \item  As with CIFAR datasets, in the absence of data augmentation, we again see a drop in performance to translation shifts in test distribution -- smaller but still significant on ConvNets and more drastic on non-convolutional architectures. In this dataset, we also see that antialiased ResNet is significantly more robust than standard ResNets. 
    \item For ConvNets, we again observe that the basic augmentation suffices to make the models robust to large translation shifts. Note that even on TinyImageNet our augmentation pipelines have translation related augmentation of at most $4$ pixels on each side ($8$ pixels in hamming distance). Thus, it is more impressive that we see generalization to translation shifts of larger magnitude than evaluated on CIFAR datasets, \ie up to $16$ pixels on each side or up to $32$ pixels in hamming distance. For non-convolutional models, the behavior is similar to CIFAR datasets -- there is improvement in robustness to translations, but the absolute test accuracies are significantly lower. 
    \item Advanced Augmentation (AA) again dramatically improves performance across all architectures and make them robust to large translations shifts. For ConvNets the performance continues to be robust even on this evaluation on a larger canvas of up $32$ pixel Hamming distance shifts. However, for non-convolutional architectures we start to see the drop in generalization performance. Notice that even this case, the models are robust up to much larger translations than present in the augmentation pipeline. In particular, cait\_xxs36 is robust up to  $20$ pixel Hamming distance shifts and resmlp\_12 up to $14$ pixels, even though the maximum magnitude of translation shifts in the augmentation pipeline is only $8$ pixels in Hamming distance. 
\end{enumerate}
In summary, our conclusions from experiments on CIFAR-10 and CIFAR-100 datasets are also supported by our results on TinyImageNet. Additionally, we see the points at which the meta generalization to larger translation shifts from translation augmentation of up to $4$ pixel random crop starts to break for non-convolutional models.

\section{Limitations}\label{sec:limitations} We study to what extent data augmentation can capture the benefits of carefully designed architectures in learning one of the fundamental priors of image processing: invariance to translations. This was a natural candidate for us as it is an instance where there is clearly a ``good" architecture, and we could focus on how augmentations might play a complementary role. To get a broader picture about learning good image priors, such studies could also be extended to other natural invariances, including other $2D$ invariances such as scale and rotations, as well as $3D$ invariances such as occlusion and lighting. Such studies would indeed be valuable, but due to combinatorial increase in number of configurations we can test, we restrict our focus to one key invariance. 

We also restrict our study to only one representative architecture from each family. In our initial experiments, we did not see significant difference in qualitative trends when using other models within these architectures, but we did not do a full study to make conclusive statements about how architectural choices like width and depth affect the robustness to translation shifts.  
Finally, from the experiments on ResMLP on CIFAR and further experiments on TinyImageNet (Section~\ref{sec:tinyimagenet}), we observe that %
we see that the prior learned from augmentation are not perfect and begin to observe differences between ConvNets and and other general purpose architectures under very large translation shifts. This suggests there might be limitations to the extent of meta-generalization from the $4$ pixel random crop holds for non-convolutional models. 

\remove{\section{Additional plots} 
\paragraph{Generalization to translation shifts on CIFAR-10} We provide summary grid of our experiments on CIFAR10 for all architectures and augmentations for more detailed perusal. 

\begin{figure}[h!]
    \centering
    \begin{subfigure}{\linewidth}
    \centering
    \includegraphics[width=0.8\linewidth]{figs/heatmapsresnet18_BN.pdf}
    \caption{\textit{resnet18\_bn}, no augmentation}
    \end{subfigure}
    \begin{subfigure}{\linewidth}
    \centering
    \includegraphics[width=0.8\linewidth]{figs/heatmapsresnet18_DA_BN.pdf}
    \caption{\textit{resnet18\_bn}, basic augmentation}
    \end{subfigure}
    \begin{subfigure}{\linewidth}
    \centering
    \includegraphics[width=0.8\linewidth]{figs/heatmapsresnet18_AAtr0_BN.pdf}
    \caption{\textit{resnet18\_bn}, advanced augmentation with no translation augmentation}
    \end{subfigure}
    \begin{subfigure}{\linewidth}
    \centering
    \includegraphics[width=0.8\linewidth]{figs/heatmapsresnet18_AA_BN.pdf}
    \caption{\textit{resnet18\_bn}, advanced augmentation}
    \end{subfigure}
\caption{\textit{resnet18\_bn}. These grids are generated with the same protocol as Figure~\ref{fig:grid-eval}. See Section~\ref{sec:eval} for details.}
\end{figure}

\begin{figure}[h!]
    \centering
    \begin{subfigure}{\linewidth}
    \centering
    \includegraphics[width=0.8\linewidth]{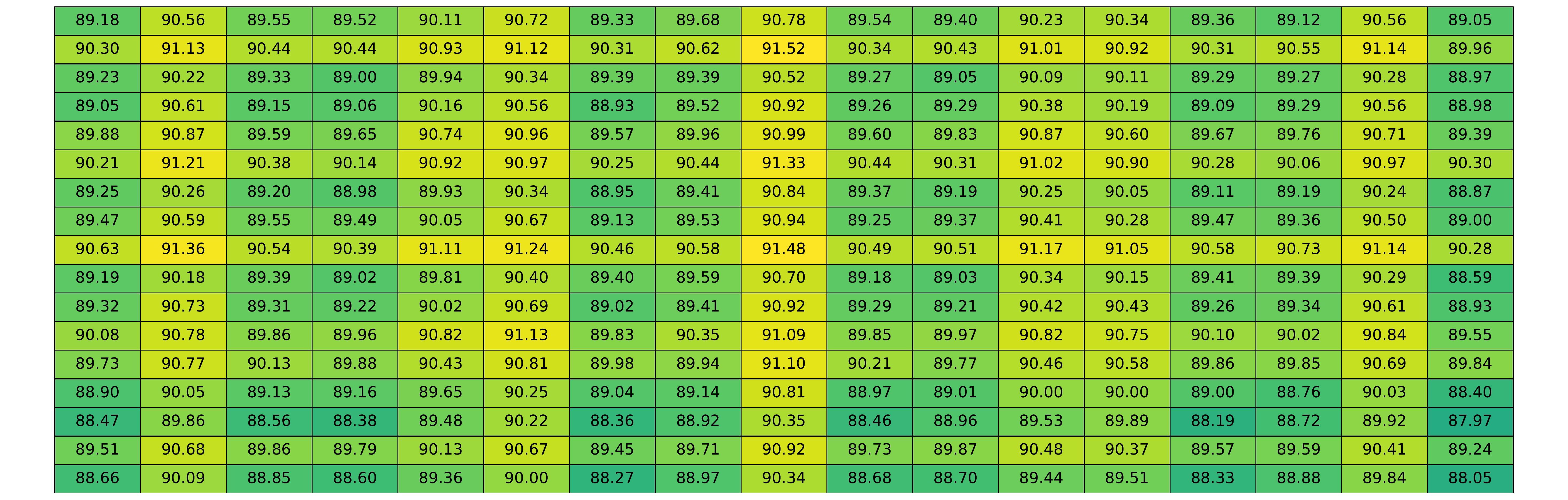}
    \caption{\textit{resnet18\_gn}, no augmentation}
    \end{subfigure}
    \begin{subfigure}{\linewidth}
    \centering
    \includegraphics[width=0.8\linewidth]{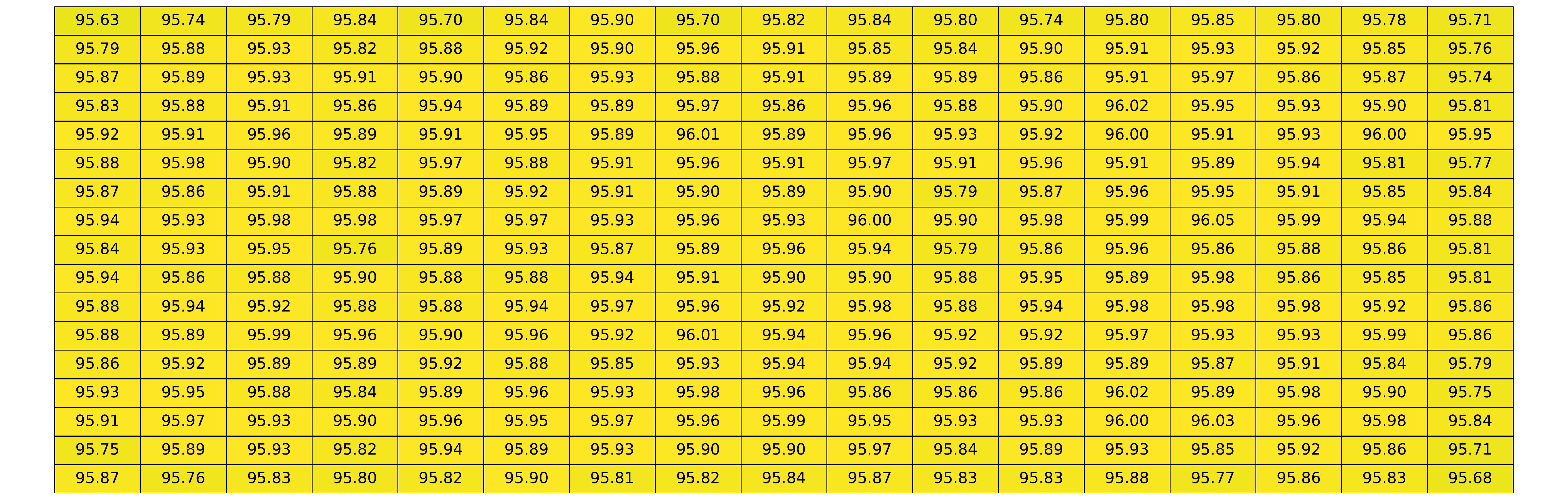}
    \caption{\textit{resnet18\_gn}, basic augmentation}
    \end{subfigure}
    \begin{subfigure}{\linewidth}
    \centering
    \includegraphics[width=0.8\linewidth]{figs/heatmapsresnet18_AAtr0_GN.pdf}
    \caption{\textit{resnet18\_gn}, advanced augmentation with no translation augmentation}
    \end{subfigure}
    \begin{subfigure}{\linewidth}
    \centering
    \includegraphics[width=0.8\linewidth]{figs/heatmapsresnet18_AA_GN.pdf}
    \caption{\textit{resnet18\_gn}, advanced augmentation}
    \end{subfigure}
\caption{\textit{resnet18\_gn}. These grids are generated with the same protocol as Figure~\ref{fig:grid-eval}. See Section~\ref{sec:eval} for details.}
\end{figure}

\begin{figure}[h!]
    \centering
    \begin{subfigure}{\linewidth}
    \centering
    \includegraphics[width=0.8\linewidth]{figs/heatmapscait_xxs36.pdf}
    \caption{\textit{cait\_xxs36}, no augmentation}
    \end{subfigure}
    \begin{subfigure}{\linewidth}
    \centering
    \includegraphics[width=0.8\linewidth]{figs/heatmapscait_xxs36_DA.pdf}
    \caption{\textit{cait\_xxs36}, basic augmentation}
    \end{subfigure}
    \begin{subfigure}{\linewidth}
    \centering
    \includegraphics[width=0.8\linewidth]{figs/heatmapscait_xxs36_AAtr0.pdf}
    \caption{\textit{cait\_xxs36}, advanced augmentation with no translation augmentation}
    \end{subfigure}
    \begin{subfigure}{\linewidth}
    \centering
    \includegraphics[width=0.8\linewidth]{figs/heatmapscait_xxs36_AA.pdf}
    \caption{\textit{cait\_xxs36}, advanced augmentation}
    \end{subfigure}
\caption{\textit{cait\_xxs36}. These grids are generated with the same protocol as Figure~\ref{fig:grid-eval}. See Section~\ref{sec:eval} for details.}
\end{figure}

\begin{figure}[h!]
    \centering
    \begin{subfigure}{\linewidth}
    \centering
    \includegraphics[width=0.8\linewidth]{figs/heatmapsvit_tiny.pdf}
    \caption{\textit{vit\_tiny}, no augmentation}
    \end{subfigure}
    \begin{subfigure}{\linewidth}
    \centering
    \includegraphics[width=0.8\linewidth]{figs/heatmapsvit_tiny_DA.pdf}
    \caption{\textit{vit\_tiny}, basic augmentation}
    \end{subfigure}
    \begin{subfigure}{\linewidth}
    \centering
    \includegraphics[width=0.8\linewidth]{figs/heatmapsvit_tiny_AAtr0.pdf}
    \caption{\textit{vit\_tiny}, advanced augmentation with no translation augmentation}
    \end{subfigure}
    \begin{subfigure}{\linewidth}
    \centering
    \includegraphics[width=0.8\linewidth]{figs/heatmapsvit_tiny_AA.pdf}
    \caption{\textit{vit\_tiny}, advanced augmentation}
    \end{subfigure}
\caption{\textit{vit\_tiny}. These grids are generated with the same protocol as Figure~\ref{fig:grid-eval}. See Section~\ref{sec:eval} for details.}
\end{figure}

\begin{figure}[h!]
    \centering
    \begin{subfigure}{\linewidth}
    \centering
    \includegraphics[width=0.8\linewidth]{figs/heatmapsresmlp_12_224.pdf}
    \caption{\textit{resmlp\_12}, no augmentation}
    \end{subfigure}
    \begin{subfigure}{\linewidth}
    \centering
    \includegraphics[width=0.8\linewidth]{figs/heatmapsresmlp_12_224_DA.pdf}
    \caption{\textit{resmlp\_12}, basic augmentation}
    \end{subfigure}
    \begin{subfigure}{\linewidth}
    \centering
    \includegraphics[width=0.8\linewidth]{figs/heatmapsresmlp_12_224_AAtr0.pdf}
    \caption{\textit{resmlp\_12}, advanced augmentation with no translation augmentation}
    \end{subfigure}
    \begin{subfigure}{\linewidth}
    \centering
    \includegraphics[width=0.8\linewidth]{figs/heatmapsresmlp_12_224_AA.pdf}
    \caption{\textit{resmlp\_12}, advanced augmentation}
    \end{subfigure}
\caption{\textit{resmlp\_12}. These grids are generated with the same protocol as Figure~\ref{fig:grid-eval}. See Section~\ref{sec:eval} for details.}
\end{figure}}
\end{document}